\documentclass[preprint,12pt]{elsarticle}

\usepackage{amssymb}
\usepackage{amsmath}
\usepackage{amsfonts}
\usepackage{amsthm}

\usepackage{xcolor, soul, framed}
\colorlet{shadecolor}{yellow}

\usepackage{graphicx}   
\graphicspath{{./}{./jpeg/}}

\usepackage[ruled,vlined,linesnumbered]{algorithm2e}
\SetKwComment{Comment}{/* }{ */}
\SetKwInOut{Input}{Input}
\SetKwInOut{Output}{Output}
\SetKwData{Data}{Persistent Data}
\SetKwFunction{FnSelectStrongest}{SelectStrongestFrom}

\usepackage{array}
\usepackage[caption=false,font=normalsize,labelfont=sf,textfont=sf]{subfig}
\usepackage{textcomp}
\usepackage{stfloats}
\usepackage{url}
\usepackage{xurl}
\usepackage{verbatim}
\usepackage{tabularx}
\usepackage{booktabs}
\usepackage{multirow}
\usepackage{rotating}
\usepackage{booktabs}
\usepackage{threeparttable} 
\usepackage[ruled,vlined,linesnumbered]{algorithm2e}
\usepackage{subcaption}
\usepackage{hyperref}
\theoremstyle{remark}

\newcommand{\Rmnum}[1]{\expandafter\@slowromancap\romannumeral #1@}

\journal{Knowledge-Based Systems}

\begin{document}

\begin{frontmatter}



\title{Governing Strategic Dynamics: Equilibrium Stabilization via Divergence-Driven Control}


\author[aff1]{Hao~Shi}
\author[aff1]{Fangfang~Xie}
\author[aff1]{Xi~Li\corref{cor1}}
\ead{lixi@aeu.edu.cn}
\cortext[cor1]{Corresponding author.}

\affiliation{organization={Shijiazhuang Campus, Army Engineering University},
            addressline={No.~97 Heping West Road, Xinhua District}, 
            city={Shijiazhuang},
            postcode={061000}, 
            state={Hebei},
            country={China}}
\begin{abstract}
Black-box coevolution in mixed-motive games is often undermined by opponent-drift non-stationarity and noisy rollouts, which distort progress signals and can induce cycling, Red-Queen dynamics, and detachment.
We propose the \emph{Marker Gene Method} (MGM), a curriculum-inspired governance mechanism that stabilizes selection by anchoring evaluation to cross-generational marker individuals, together with DWAM and conservative marker-update rules to reduce spurious updates.
We also introduce NGD-Div, which adapts the key update threshold using a divergence proxy and natural-gradient optimization.
We provide theoretical analysis in strictly competitive settings and evaluate MGM integrated with evolution strategies (MGM-E-NES) on coordination games and a resource-depletion Markov game.
MGM-E-NES reliably recovers target coordination in Stag Hunt and Battle of the Sexes, achieving final cooperation probabilities close to $(1,1)$ (e.g., $0.991\pm0.01/1.00\pm0.00$ and $0.97\pm0.00/0.97\pm0.00$ for the two players).
In the Markov resource game, it maintains high and stable state-conditioned cooperation across 30 seeds, with final cooperation of $\approx 0.954/0.980/0.916$ in \textsc{Rich}/\textsc{Poor}/\textsc{Collapsed} (both players; small standard deviations), indicating welfare-aligned and state-dependent behavior.
Overall, MGM-E-NES transfers across tasks with minimal hyperparameter changes and yields consistently stable training dynamics, showing that top-level governance can substantially improve the robustness of black-box coevolution in dynamic environments.
\end{abstract}


\begin{keyword}
\texorpdfstring{Coevolution \sep Mixed-motive games \sep Markov games \sep Black-box optimization \sep Natural Evolution Strategies \sep Non-stationary evaluation \sep Robust learning dynamics \sep Equilibrium stabilization}{Coevolution, Mixed-motive games, Markov games, Black-box optimization, Natural Evolution Strategies, Non-stationary evaluation, Robust learning dynamics, Equilibrium stabilization}
\end{keyword}

\end{frontmatter}




\section{Introduction}

A central difficulty in co-evolutionary learning is the non-stationarity of evaluation: an individual’s measured fitness---and even its \emph{relative ranking}---can change drastically as the opponent population drifts.
This makes selection pressure unreliable, often leading to cycling, forgetting, or training collapse, especially in non-transitive or multi-equilibrium games where within-generation ``progress'' can be misleading.

We address this by introducing the \emph{Marker Gene Method (MGM)}, a core control mechanism for mixed-motive coevolution inspired by curriculum learning.
Our key insight is that many coevolutionary pathologies (e.g., intransitive cycles, Red Queen dynamics, and detachment) stem from the lack of a stable yet continually improving evaluation reference: as the opponent population drifts, the fitness scale itself changes, making selection signals unreliable.
MGM therefore \emph{anchors} evaluation with a \emph{marker gene} that is assessed every generation as a persistent reference, while avoiding stagnation by \emph{adaptively updating} this anchor.
Concretely, MGM couples (i) a conservative marker-gene update mechanism with (ii) a dynamic-weighting criterion (DWAM) that regulates when the anchor should advance, balancing specialization against the current marker with generalization against the opponent population—effectively implementing a curriculum of “specialize, generalize, then update”.
For the NES instantiation studied in this paper, we additionally maintain a bounded FIFO archive of past markers (size $H$) to support batched evaluation and bookkeeping under stochastic rollouts; the MGM core update commits to a new marker once the update criteria are met rather than relying on rollback as a primary control signal.
Together, these components turn cross-generational progress into a more reliable signal under opponent drift, improving the stability of selection in black-box coevolution.

However, a practical obstacle remains: a key MGM hyperparameter---the DWAM threshold $l$---depends on game scale and equilibrium structure, which are typically unknown \emph{a priori} and may vary across tasks.
To reduce manual tuning, we propose \emph{NGD-Div}, a lightweight governance controller that adapts $l$ online using a computable proxy signal.
NGD-Div does not estimate an exact game-theoretic quantity; instead, it regulates a single scalar to keep learning in a stable regime.

We package MGM together with NGD-Div as \emph{MGM-E}, a modular governance framework that can be instantiated with different evolutionary actuators.
In this work we adopt Natural Evolution Strategies (NES) as a differentiability-free actuator for concreteness; importantly, MGM and NGD-Div are actuator-agnostic components.

Empirically, MGM-E yields more stable learning dynamics across representative non-transitive games and coordination settings, and often improves solution quality in Markov games, including more reliable selection of Pareto-efficient equilibria in coordination benchmarks and higher cooperation rates in mixed-motive environments.
We further compare against representative alternatives such as PSRO and stabilized gradient-based methods (e.g., OGDA), highlighting complementary strengths and failure modes.

\paragraph{Contributions}
\begin{itemize}
\item We propose MGM, a marker-gene-based cross-generational anchoring mechanism with a bounded-archive rollback rule for stabilizing co-evolutionary evaluation.
\item We provide a theoretical analysis of the MGM core mechanism. Under bounded payoffs, finite-sample evaluation, and weak-dependence assumptions, we establish a separation-of-time-scales result: entering a Nash-equilibrium neighborhood occurs on a much shorter time scale than leaving it, and under small stochastic perturbations the expected escape time from the neighborhood grows exponentially in the inverse perturbation scale.
\item We propose NGD-Div, a governance controller that adaptively tunes the DWAM threshold $l$ via a proxy-driven natural-gradient update, adding only a constant-factor overhead on top of the underlying actuator.
\item We instantiate the framework with NES (MGM-E-NES) and validate it on non-transitive, coordination, and Markov games, including extensive ablations and comparisons against PSRO/OGDA-style baselines.
\end{itemize}

\section{Related Works and Backgrounds}
\label{sec:II}

This section reviews two complementary perspectives.
We first summarize three well-documented pathologies in population-based coevolution that jointly hinder stable learning dynamics.
We then survey major algorithmic lines of work for learning in games and multi-agent settings, highlighting their typical assumptions and limitations in mixed-motive and non-transitive environments.

\subsection{Pathologies in Population Coevolution}
\label{ssec:pathologies}

\subsubsection{Intransitivity}
Intransitivity is prevalent in population coevolution.
Jong characterizes intransitivity via a relation $R$ in which $aRb \land bRc$ does not imply $aRc$~\cite{Jong_2004}.
In coevolutionary settings, such non-transitive dominance often manifests as rock--paper--scissors-like cycles, which can trap the search process in persistent oscillations and impede convergence to stable solutions~\cite{Yao2025Pathologies}.
At the same time, intransitivity can sometimes be beneficial by maintaining diversity and preventing premature concentration on a single strategy~\cite{Richter2015Coevolutionary}.
Fully eliminating intransitivity is difficult, especially in games with more than three strategies and in high-dimensional strategy spaces~\cite{Nolfi1998Coevolving,Szolnoki2014Cyclic}.

From an evolutionary game theory viewpoint, under strictly competitive interactions and without dissipative terms, intransitive systems can exhibit Hamiltonian-like behavior with heteroclinic cycles, preventing volume contraction in the strategy-frequency simplex and leading to persistent cycling~\cite{Szolnoki2014Cyclic}.
Consequently, population states may drift along the simplex rather than contracting toward a (mixed) Nash equilibrium; the classic rock--paper--scissors model is a canonical example~\cite{Yao2025Pathologies}.
With finite populations and stochastic mutation, dynamics deviate from ideal Hamiltonian systems, yet cyclic phenomena often persist as damped or sustained oscillations~\cite{Perdigao2023Complexity}.
Such oscillations can distort fitness evaluation: strong strategies may be misjudged as weak when assessed at unfavorable phases of the cycle, thereby biasing selection and slowing progress~\cite{Nolfi2025Global}.

A broad class of coevolutionary algorithms therefore introduces additional mechanisms to cope with intransitivity.
Examples include runtime-oriented approaches that explicitly archive equilibrium-like solutions to prevent ``forgetting''~\cite{Fajardo2023Runtime}, distribution-estimation methods whose success depends on reliable frequency estimates under sufficient diversity~\cite{Benford2024Runtime}, and methods that impose explicit structural information or preference protocols to enforce a more stable search direction~\cite{Fajardo2024Ranking,Lai2021Solving}.
Hall-of-Fame style archives are widely used to preserve elites, though their long-term influence can extend over many generations and may not be fully captured by purely current-fitness-based criteria~\cite{Zychowski2024Coevolutionary}.

\subsubsection{Red Queen effect}
Coevolutionary dynamics are often described by the Red Queen effect: interacting populations must continually adapt because each population's progress changes the fitness landscape faced by the other~\cite{Solé_2022}.
This phenomenon is double-edged.
On one hand, sustained competitive pressure can maintain diversity and mitigate premature convergence, which underlies its widespread use in evolutionary modeling across ecology and economics~\cite{Antero2022Sourcing,Lamsdell2021Conquest,Robson2005Complex}.
On the other hand, it can trap populations in unproductive cycles of ``running to stay in the same place''~\cite{Cliff1995Tracking}, where improvements are transient and oscillatory rather than cumulative~\cite{Sole2014Red,Rabajante2015Red}.

From a dynamical systems  perspective, Red Queen dynamics can lead to persistent oscillations or even chaos attractors.
Extensions of Lotka--Volterra-style models illustrate that mutual adaptation can sustain population oscillations instead of convergence to stable equilibrium points~\cite{Dercole2010Chaotic,Marrow1992Coevolution}.
Comparative models suggest that genetic drift may weaken but not necessarily eliminate these oscillations~\cite{Schenk2020How}.
Algorithmically, several approaches aim to stabilize the evolutionary direction using additional structure, such as multi-objective formulations with preference articulation~\cite{Zychowski2018Addressing}, heuristic operators that leverage historical performance~\cite{Li2024Competitive}, or knowledge-guided mechanisms that incorporate prior domain information~\cite{Zhou2024KnowledgeGuided}.
These methods can be effective but typically rely on the availability and quality of auxiliary signals (preference rules, memory relevance, or domain knowledge), which may be fragile under strong intransitivity or rapid landscape shifts.

\subsubsection{Detachment}
Detachment refers to a regime in which one population becomes sufficiently advantaged that its opponent experiences a flattened fitness landscape, reducing directional selection and allowing drift to dominate~\cite{Cartlidge2004Combating}.
This can compromise coevolution by effectively turning one side into a near-static objective, weakening the interactive benefits of coevolution~\cite{Antonio2018Coevolutionary}.
Two commonly cited triggers are (i) vanishing fitness gradients that erode selective pressure and induce random-walk behavior~\cite{Dieckmann1996Dynamical,Rikvold2009Complex,Dercole2006Coevolution}, and (ii) mismatched evolutionary rates that decouple the populations and introduce evolutionary lag~\cite{Dercole2006Coevolution}.
Such decoupling has also been observed empirically in long-horizon evolutionary experiments~\cite{Marshall2022Longterm}.

Algorithmically, detachment can cause a CCEA to degenerate into a single-population optimizer, undermining multi-agent trade-offs~\cite{Ma2019Survey}.
Even when interactions are not fully severed, residual competition can still induce oscillatory behavior in partially decoupled systems~\cite{Kim2013Coevolution}.
Stabilization strategies often overlap with those used for Red Queen dynamics, e.g., alternating evolution schemes and other protocols that prevent either population from escaping selective pressure indefinitely, albeit with additional parameters that can be sensitive to tuning~\cite{Zychowski2024Coevolutionary}.

\subsection{Algorithmic Lines of Work for Learning in Games}
\label{ssec:lines}

Beyond coevolutionary heuristics, a large literature studies learning dynamics in games through different algorithmic lenses.

\paragraph{Gradient-based dynamics}
One line of inquiry relies on gradient information.
Methods such as Optimistic Gradient Descent (OGDA)~\cite{daskalakis2018limit,boct2025fast} and opponent-shaping algorithms such as LOLA~\cite{LOLA,Kopacz_2023_09} can exhibit rapid convergence in structured settings (notably two-player zero-sum games).
However, in mixed-motive environments they may cycle, converge to Pareto-suboptimal equilibria, or become highly sensitive to initialization when multiple Nash equilibria exist.
They also commonly rely on differentiability and accurate gradient estimation, assumptions that can be restrictive in simulator-based or black-box settings.

\paragraph{Meta-game and oracle-based methods}
Policy-Space Response Oracles (PSRO) and variants~\cite{NIPS2017_3323fe11,Tang_2025} iteratively expand a policy set by computing approximate best responses and solving a meta-game over discovered strategies.
While PSRO-style methods are expressive in principle, they can incur substantial expansion and evaluation costs as the strategy set grows, and practical guarantees for unified behavior across both zero-sum and mixed-motive settings remain limited.

\paragraph{Equilibrium selection and evolutionary dynamics}
Classic Fictitious Play~\cite{brown1951fictitious,Lazarsfeld_2025} and Replicator Dynamics~\cite{TAYLOR1978145,Gao_2024_04} form foundational baselines linking learning to equilibrium concepts, but both can exhibit persistent cycles in non-transitive games~\cite{Ju_2024}.
Moreover, their behavior can be affected by Red Queen dynamics and population decoupling effects in practical coevolutionary regimes~\cite{Yao_Chong_2025}.

\paragraph{Regret minimization}
Counterfactual Regret Minimization (CFR)~\cite{zinkevich2007regret,Sychrovsky_2025} provides strong guarantees in two-player zero-sum imperfect-information games, yet such guarantees do not directly extend to general-sum mixed-motive interactions.

\subsection{Summary}
\label{ssec:rw_summary}
In summary, prior work addresses instability in games either by refining agents' adaptation dynamics (e.g., gradient-based updates), by adding external scaffolding (e.g., meta-game expansion and archives), or by leveraging structural assumptions (e.g., zero-sum or extensive-form constraints).
However, mixed-motive coevolutionary settings can simultaneously exhibit intransitivity, Red Queen dynamics, and detachment, making stability and equilibrium selection challenging under non-stationary and noisy evaluation.
These observations motivate approaches that explicitly regulate coevolutionary dynamics while remaining compatible with black-box policy search.

\begin{table*}[t]
    \centering
    \caption{Comprehensive performance comparison across competitive and coordination games. For RPS games, we report both the final KL divergence and the total KL Reduction (Final - Initial) to measure final precision and convergence capability, respectively (lower is better for Final KL, more negative is better for Reduction). For coordination games, we report the final strategy profiles (mean $\pm$ std over 30 runs). Our MGM-E-NES is the only method that demonstrates both superior final precision and powerful convergence dynamics across all domains.}
    \label{tab:ultimate_judgement}
    \resizebox{\textwidth}{!}{%
    \begin{tabular}{lcccccc}
        \toprule
        \multirow{2}{*}{\textbf{Algorithm}} & \multicolumn{3}{c}{\textbf{Competitive Games (RPS)}} & \multicolumn{2}{c}{\textbf{Coordination Games (Final Strategy Profile)}} \\
        \cmidrule(lr){2-4} \cmidrule(lr){5-6}
        & 3D RPS & 100D RPS & 1000D RPS & Stag Hunt (Target: (1,1)) & Battle of Sexes (Target: (1,1)) \\
        \midrule
        \multicolumn{1}{l}{\textit{Eval. Budget}} & \textit{3.2e4} & \textit{1.2e7} & \textit{2e7} & \textit{3.2e4} & \textit{3.2e4}\\
        \multicolumn{1}{l}{\textit{Final Target $l$ (Ours)}} & \textit{-0.000} & \textit{-0.000} & \textit{-0.000} & \textit{3.391} & \textit{1.100} \\ 
        \midrule
        \multirow{2}{*}{\textbf{MGM-E-NES (Ours)}} 
        & \begin{tabular}{@{}l@{}}Final: 5.10e-4 $\pm$ 1.78e-7\\ Reduc: -7.81e-2 $\pm$ 3.05e-3\end{tabular}
        & \begin{tabular}{@{}l@{}}Final: \textbf{4.32e-3 $\pm$ 7.35e-4}\\ Reduc: -6.98e-3 $\pm$ 7.55e-4\end{tabular} 
        & \begin{tabular}{@{}l@{}}Final: 1.08e-2 $\pm$ 2.75e-4\\ Reduc: -4.38e-4 $\pm$ 1.07e-4\end{tabular}
        & \begin{tabular}{@{}l@{}}P1: 0.99 $\pm$ 0.01\\ P2: 1.00 $\pm$ 0.00\end{tabular} 
        & \begin{tabular}{@{}l@{}}P1: 0.97 $\pm$ 0.00\\ P2: 0.97 $\pm$ 0.00\end{tabular} \\ 
        \midrule
        \multirow{2}{*}{OGDA} 
        & \begin{tabular}{@{}l@{}}Final: \textbf{5.42e-6 $\pm$ 6.45e-11} \\ Reduc: -7.54e-2 $\pm$ 2.85e-3\end{tabular}
        & \begin{tabular}{@{}l@{}}Final: 1.13e-2 $\pm$ 1.12e-3\\ Reduc: +2.41e-4 $\pm$ 4.75e-4\end{tabular}
        & \begin{tabular}{@{}l@{}}Final: 1.12e-2 $\pm$ 3.69e-4\\ Reduc: -8.65e-7 $\pm$ 7.88e-6\end{tabular}
        & \begin{tabular}{@{}l@{}}P1: 0.60 $\pm$ 0.49 \\ P2: 0.60 $\pm$ 0.49\end{tabular}
        & \begin{tabular}{@{}l@{}}P1: 0.60 $\pm$ 0.49 \\ P2: 0.60 $\pm$ 0.49\end{tabular} \\ 
        \midrule
        \multirow{2}{*}{PureNES} 
        & \begin{tabular}{@{}l@{}}Final: 8.82e-2 $\pm$ 4.28e-3 \\ Reduc: +9.58e-3 $\pm$ 5.62e-4\end{tabular}
        & \begin{tabular}{@{}l@{}}Final: 1.22e-2 $\pm$ 1.19e-3\\ Reduc: +8.96e-4 $\pm$ 5.25e-4\end{tabular}
        & \begin{tabular}{@{}l@{}}Final: 1.14e-2 $\pm$ 6.47e-8\\ Reduc: +1.62e-4 $\pm$ 7.85e-9\end{tabular}
        & \begin{tabular}{@{}l@{}}P1: 0.99 $\pm$ 0.01\\ P2: 1.00 $\pm$ 0.00\end{tabular}
        & \begin{tabular}{@{}l@{}}P1: \textbf{0.99 $\pm$ 0.00}\\ P2: \textbf{0.99 $\pm$ 0.00}\end{tabular} \\ 
        \midrule
        \multirow{2}{*}{LOLA} 
        & \begin{tabular}{@{}l@{}}Final: 8.62e-2 $\pm$ 4.60e-3 \\ Reduc: +1.08e-2 $\pm$ 5.74e-4\end{tabular}
        & \begin{tabular}{@{}l@{}}Final: 1.13e-2 $\pm$ 1.16e-3\\ Reduc: +2.06e-4 $\pm$ 2.83e-4\end{tabular}
        & \begin{tabular}{@{}l@{}}Final: 1.12e-2 $\pm$ 3.67e-4\\ Reduc: +6.06e-7 $\pm$ 2.50e-6\end{tabular}
        & \begin{tabular}{@{}l@{}}P1: 0.47 $\pm$ 0.50 \\ P2: 0.53 $\pm$ 0.50\end{tabular}
        & \begin{tabular}{@{}l@{}}P1: 0.00 $\pm$ 0.00 \\ P2: 1.00 $\pm$ 0.00\end{tabular} \\ 
        \midrule
        \multirow{2}{*}{FP} 
        & \begin{tabular}{@{}l@{}}Final: 1.88e-5 $\pm$ 3.74e-12\\ Reduc: +1.88e-5 $\pm$ 3.74e-12\end{tabular}
        & \begin{tabular}{@{}l@{}}Final: 4.59e+0 $\pm$ 2.06e-4\\ Reduc: +4.59e+0 $\pm$ 2.06e-4\end{tabular}
        & \begin{tabular}{@{}l@{}}Final: 5.97e+0 $\pm$ 1.09e-3\\ Reduc: +5.97e+0 $\pm$ 1.09e-3\end{tabular}
        & \begin{tabular}{@{}l@{}}P1:  0.47$\pm$ 0.50\\ P2: 0.47$\pm$ 0.50\end{tabular}
        & \begin{tabular}{@{}l@{}}P1: 0.47 $\pm$ 0.50 \\ P2: 0.47 $\pm$ 0.50\end{tabular} \\ 
        \midrule
        \multirow{2}{*}{PSRO} 
        & \begin{tabular}{@{}l@{}}Final: 7.09e-4 $\pm$ 3.51e-7 \\ Reduc: \textbf{-1.39e-1 $\pm$ 5.41e-2}\end{tabular}
        & \begin{tabular}{@{}l@{}}Final: 1.17e-2 $\pm$ 2.54e-3\\ Reduc: \textbf{-2.65e-2 $\pm$ 4.87e-2}\end{tabular}
        & \begin{tabular}{@{}l@{}}Final: \textbf{1.06e-3 $\pm$ 1.13e-4}\\ Reduc: \textbf{-1.20e-2 $\pm$ 1.70e-2}\end{tabular}
        & \begin{tabular}{@{}l@{}}P1: 0.23 $\pm$ 0.42 \\ P2: 0.23 $\pm$ 0.42\end{tabular}
        & \begin{tabular}{@{}l@{}}P1: 0.67 $\pm$ 0.47 \\ P2: 0.67 $\pm$ 0.47\end{tabular} \\ 
        \midrule
        \multirow{2}{*}{DPG} 
        & \begin{tabular}{@{}l@{}}Final: 1.10e-2 $\pm$ 4.19e-5 \\ Reduc: 0.00e+0 $\pm$ 0.00\end{tabular}
        & \begin{tabular}{@{}l@{}}Final: 1.40e-2 $\pm$ 2.01e-3\\ Reduc: 0.00e+0 $\pm$ 0.00e+0\end{tabular}
        & \begin{tabular}{@{}l@{}}Final: 1.40e-2 $\pm$ 1.86e-3\\ Reduc: 0.00e+0 $\pm$ 0.00\end{tabular}
        & \begin{tabular}{@{}l@{}}P1: \textbf{1.00 $\pm$ 0.00} \\ P2: \textbf{1.00 $\pm$ 0.00}\end{tabular}
        
        &\begin{tabular}{@{}l@{}}P1: 0.50 $\pm$ 0.50 \\ P2: 0.50 $\pm$ 0.50\end{tabular} \\ 
        \bottomrule
    \end{tabular}
    }
\end{table*}

\section{
The Governance Framework
}
\label{sec:III}
\subsection{Problem Formulation}
Establishing convergence guarantees for co-evolutionary algorithms in general-sum games is notoriously difficult due to non-stationarity and cyclic strategic responses.
To obtain an analytically tractable yet representative setting for studying strategic cycling (e.g., RPS-type dynamics), we conduct our theoretical analysis in the class of \emph{symmetric strictly competitive games (SCGs)}.

\paragraph{Symmetric finite game}
Let $S$ be a finite action set with $|S|=d$.
Both players use mixed strategies $x,y \in \Delta_{d-1}$, and player~1's expected payoff is
\begin{equation}
U(x,y) = x^\top A y,
\end{equation}
for some payoff matrix $A\in\mathbb{R}^{d\times d}$.

\paragraph{Strictly competitive assumption and normalization}
A game is strictly competitive if the players' preferences are perfectly opposed: for any two outcomes $o,o'$,
\begin{equation}
u_1(o) > u_1(o') \iff u_2(o) < u_2(o').
\end{equation}
Under this condition, the game is \emph{order-equivalent} to a zero-sum game: there exist strictly increasing transformations of utilities such that the transformed payoffs satisfy $u_1'(o) + u_2'(o)=0$.
In the symmetric setting, this yields a normalized anti-symmetric representation
\begin{equation}
R = -R^\top,\qquad U(x,y)=x^\top R y,
\end{equation}
where we additionally apply a bounded normalization (e.g., mapping payoffs to $[-1,1]$) to standardize scale across tasks.

\paragraph{Remark (role of monotone transforms)}
The monotone transform and normalization are introduced for analytical convenience and consistent visualization; they preserve strategic ordering and Nash equilibria structure up to order-equivalence.
In the proofs, the induced Lipschitz constant of the transformation appears only as a multiplicative constant in intermediate bounds and does not change the qualitative stability conclusions.

\paragraph{Goal}
We analyze the strategic dynamics on the product simplex $\Delta_{d-1}\times\Delta_{d-1}$ and aim to design a governance layer that stabilizes the induced learning dynamics (i.e., mitigates persistent cycling and improves equilibrium-seeking behavior) without per-game tuning.
This motivates the divergence-based control signals used in later sections.

\paragraph{Payoff normalization (analysis/visualization only)}
For theoretical analysis and visualization purposes, we apply a monotone affine transformation to rescale the raw payoff to the unit interval.
Given a scalar payoff $u\in[u_{\min},u_{\max}]$, we define
\begin{equation}
\bar{u} = \frac{u-u_{\min}}{u_{\max}-u_{\min}+\varepsilon}\in[0,1],
\label{eq:payoff_norm_01}
\end{equation}
where $\varepsilon>0$ avoids division by zero when $u_{\max}=u_{\min}$.
This normalization is introduced purely to standardize the plotting/analysis scale (e.g., for visualizing DWAM surfaces) and does not affect any qualitative conclusions, since it is a monotone transformation that preserves payoff ordering within each game.
Unless otherwise stated, all experiments and applications use the original (unnormalized) payoff $u$.

\noindent
In particular, we will leverage the divergence of the induced vector field as a diagnostic signal of local rotational behavior, and use it to construct a closed-loop stabilizing update in our governance layer.
\subsection{
The Marker Gene Governance Layer
 }
To counteract the instability inherent in Competitive Co-evolutionary Algorithms (CCEAs), such as intransitivity and the Red Queen effect, we propose the Marker Gene Method (MGM). MGM provides a stable evolutionary gradient by evaluating individuals against two distinct criteria. Each population maintains a long-term Marker Gene that serves as a stable benchmark for solution quality. The fitness of an individual is then calculated as a weighted sum of: 1) Base Fitness, its performance against this stable marker gene, and 2) Generalize Fitness, its average performance against the current opponent population. This dual-fitness evaluation steers evolution towards robust solutions, avoiding the pitfalls of relying solely on transient, snapshot-based fitness. 

MGM is operationalized through two primary mechanisms:
\begin{enumerate}
\item \textbf{A Dynamic Weight Adjustment Mechanism} to balance the influence of Base and Generalize fitness. 
\item \textbf{A Marker Gene Update Mechanism} to ensure the benchmark’s stability and relevance. 
\end{enumerate}

Before introducing the governance modules, we specify the evaluation protocol used to estimate fitness under a fixed rollout budget, as it determines both the optimization signal and the computational cost.

At each generation $t$, each candidate policy/individual $x_i^{(t)}$ is evaluated by interacting with a randomly sampled subset of opponents from the opposing population.
Let the opposing population at generation $t$ be $\mathcal{P}_{-i}^{(t)}$ with size $N_{-i}^{(t)} \!=\! |\mathcal{P}_{-i}^{(t)}|$.
For each $x_i^{(t)}$, we sample a subset $\mathcal{S}_i^{(t)} \subset \mathcal{P}_{-i}^{(t)}$ \emph{uniformly without replacement} with
\begin{equation}
\label{eq:k_sampling_ratio}
|\mathcal{S}_i^{(t)}| \;=\; k_{-i}^{(t)} \;=\; \left\lceil \rho\, N_{-i}^{(t)} \right\rceil,\qquad \rho\in(0,1].
\end{equation}
The generalization score is then estimated as
\begin{equation}
\label{eq:generalization_estimator}
\widehat{G}_i^{(t)} \;=\; \frac{1}{k_{-i}^{(t)}} \sum_{y \in \mathcal{S}_i^{(t)}} \bar{u}\!\left(x_i^{(t)},y\right),
\end{equation}
where $\bar{u}\in[0,1]$ denotes the normalized payoff used only for theoretical analysis and visualization (Eq.~\ref{eq:payoff_norm_01}).
We resample $\mathcal{S}_i^{(t)}$ independently across evaluation rounds; random resampling reduces persistent pairing effects and supports concentration arguments under weak dependence (see Supplementary Material).
This protocol yields an unbiased estimator of the full-population mean payoff while keeping the evaluation cost per generation proportional to $N_i^{(t)} \cdot k_{-i}^{(t)} = O\!\bigl(\rho\,N_i^{(t)}N_{-i}^{(t)}\bigr)$.

\addtolength{\textfloatsep}{-10pt} 

\begin{figure}[htbp] 
  \centering
  \includegraphics[width=0.9\columnwidth]{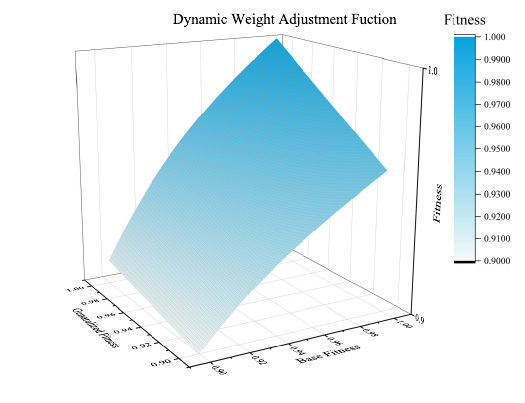}
  \caption{This figure provides a local visualization of the Dynamic Weight Adjustment Mechanism (DWAM) in the vicinity of the threshold. Specifically, the 3D surface shows the resulting comprehensive fitness $\widehat{F}_i^{(t)}$ when the marker-based score $\widehat{B}_i^{(t)}$ and the generalization score $\widehat{G}_i^{(t)}$ vary within $[0.9,1.0]$, while the threshold is fixed at $l=0.9$ and the scale parameter is fixed at $s=10^{2}$.}
  \label{3d-weights}
\end{figure}

\subsubsection{Dynamic Weight Adjustment Mechanism (DWAM)}
\label{sec:dynamic_weight_adj}

DWAM adaptively balances two evaluation criteria for each individual: (i) \emph{base fitness} against the current marker gene, encouraging stable progress along a reference direction; and (ii) \emph{generalize fitness} against a uniformly sampled subset of the current opponent population, discouraging overfitting to the marker and improving robustness.

\paragraph{Base and generalize fitness}
Let $x_i^{(t)}$ be an individual from population $i$ at generation $t$, and let $M_{-i}^{(t)}$ denote the opponent marker gene maintained by the Marker Archive. Using the normalized payoff $\tilde{u}\in [0,1]$ (Eq.~\ref{eq:payoff_norm_01}), we define
\begin{align}
\widehat{B}_i^{(t)} &= \tilde{u}\!\left(x_i^{(t)},\, M_{-i}^{(t)}\right),
\label{eq:base_fitness}\\
\widehat{G}_i^{(t)} &= \frac{1}{m}\sum_{y\in \mathcal{S}_i^{(t)}} \tilde{u}\!\left(x_i^{(t)},\, y\right),
\qquad \mathcal{S}_i^{(t)} \subset \mathcal{P}_{-i}^{(t)},\ |\mathcal{S}_i^{(t)}|=k_{-i}^{(t)}=\lceil \rho N_{-i}^{(t)}\rceil ,
\label{eq:gen_fitness}
\end{align}
where $\mathcal{S}_i^{(t)}$ is drawn uniformly from the current opponent population $\mathcal{P}_{-i}^{(t)}$ and resampled independently for each evaluation round.

\paragraph{Dynamic weight}
DWAM uses a target level $l$ to determine when to relax marker-centric learning and incorporate population-level generalization.
In the full method, $l$ is automatically adjusted by NGD-Div (Sec.~\ref{sec:ngd_div}); in ablations it can be fixed.

Define the deviations from the target level
\begin{align}
\beta &\,=\, \widehat{B}_i^{(t)} - l, \\
\sigma &\,=\, \widehat{G}_i^{(t)} - l,
\end{align}
and the \emph{generalization deficit}
\begin{equation}
\delta \,=\, \max(0,-\sigma).
\end{equation}
The marker weight $\alpha\in[0,1]$ is computed as
\begin{equation}
\alpha(\beta,\sigma)=
\begin{cases}
    \omega, & \beta < 0, \\[2mm]
    \omega - (\omega - 0.5)\left(1 - e^{-s\,\delta\,\beta}\right), & \beta \ge 0,
\end{cases}
\label{eq:dwam_alpha}
\end{equation}
where $\omega\in(0,1)$ is the base anchoring weight and $s>0$ controls the transition sharpness.
Following the stability condition derived in Supplementary Remark~F.6, we set $\omega=0.9$ by default.

\paragraph{Composite fitness}
DWAM outputs the composite fitness used by the evolutionary operators:
\begin{equation}
\widehat{F}_i^{(t)} \;=\; \alpha\,\widehat{B}_i^{(t)} \;+\; (1-\alpha)\,\widehat{G}_i^{(t)}.
\label{eq:composite_fitness}
\end{equation}

\paragraph{Complexity}
Let the two populations at generation $t$ have sizes $N_i^{(t)}$ and $N_{-i}^{(t)}$.
Per generation, each individual plays (i) one match against the current marker and (ii) $k_{-i}^{(t)}=\lceil \rho N_{-i}^{(t)}\rceil$ matches against uniformly sampled opponents (without replacement).
Therefore, the evaluation cost per generation is
\[
O\!\Bigl(N_i^{(t)}\bigl(1+k_{-i}^{(t)}\bigr)\Bigr)
\;=\;
O\!\bigl(\rho\,N_i^{(t)}N_{-i}^{(t)}\bigr)
\]
pairwise payoff queries (or $O(\rho\,N_i^{(t)}N_{-i}^{(t)}H)$ environment steps for horizon $H$).

\paragraph{Interpretation of the weighting rule}
DWAM induces an adaptive schedule controlled by the target level $l\in(0,1)$.

\textbf{(i) Marker-anchored regime.}
When the marker-based score is below target, $\widehat{B}_i^{(t)}<l$ (i.e., $\beta<0$), DWAM sets $\alpha=\omega$, so $\widehat{F}_i^{(t)}$ is dominated by performance against the marker. This reduces directional volatility during early-stage population instability.

\textbf{(ii) Generalization-increasing regime.}
Once $\widehat{B}_i^{(t)}\ge l$ (i.e., $\beta\ge 0$), DWAM increases the relative influence of the generalization term by decreasing $\alpha$ from $\omega$ toward $0.5$.
If generalization is below target ($\widehat{G}_i^{(t)}<l$, i.e., $\delta>0$), then $\alpha$ decays faster as $\beta$ grows, thereby down-weighting individuals that perform well only against the marker but poorly against sampled opponents (indicative of overfitting to the marker).
As generalization improves ($\widehat{G}_i^{(t)} \uparrow l$), $\delta$ decreases, slowing the decay of $\alpha$ and preventing excessive sensitivity to the stochastic generalization estimate.
When $\widehat{G}_i^{(t)}\ge l$ (i.e., $\delta=0$), we have $e^{-s\,\delta\,\beta}=1$ and thus $\alpha=\omega$, yielding a stable weighting behavior.

\paragraph{Implementation note (analysis/visualization scale)}
When payoffs are normalized to $(0,1)$ for analysis/visualization, we use $s=O(10^2)$ by default so that $\alpha$ varies smoothly yet responsively near the target level; see Fig.~\ref{3d-weights} and the sensitivity study in the supplementary material.

\subsubsection{Marker Gene Update Mechanism}
\label{sec:marker_gene_up}

The marker gene is a reference opponent selected from the opposing population, serving as a stable evolutionary anchor that mitigates directional oscillations caused by non-transitivity and Red-Queen dynamics.
Since both \emph{when} to update the marker and \emph{which} opponent to select directly affect stability and progress, MGM uses a two-stage update rule: (i) an update \emph{timing} gate (BufferCount) and (ii) a \emph{candidate} persistence filter (KeepCount), followed by a strength-based tie-break.
\begin{figure}[htbp]
  \centering
  \includegraphics[width=0.9\columnwidth]{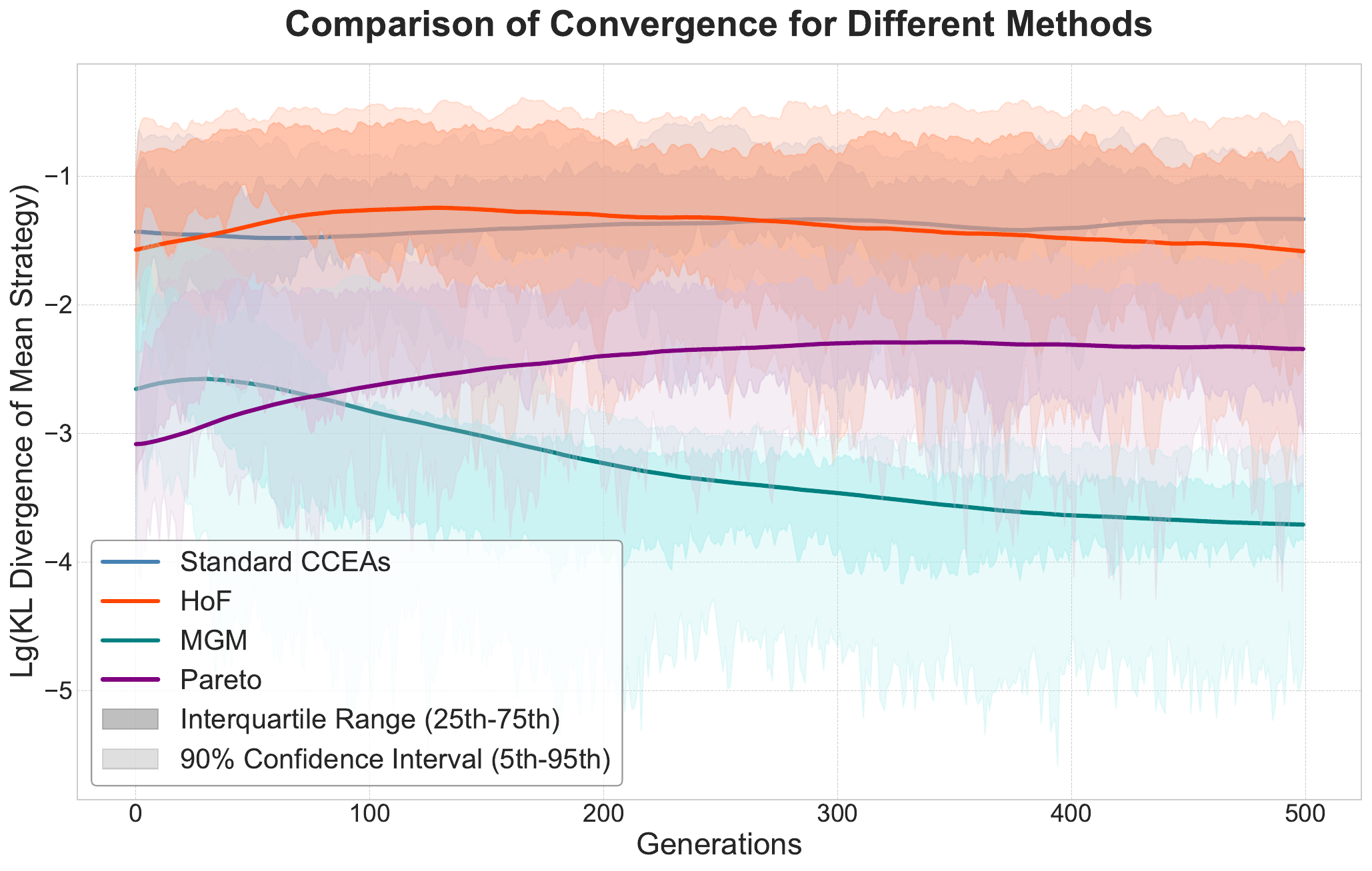}
  \caption{Convergence to the Nash equilibrium (NE) in RPS, measured by $\log_{10}$ KL divergence between the mean population strategy and the NE mixed strategy. Lower is better. Shaded bands indicate the interquartile range (25th--75th percentiles) and the empirical 5th--95th percentile interval across runs.}
  \label{comaprison_base}
\end{figure}
\paragraph{Setup}
Consider a two-population strictly competitive setting with populations $A^{(t)}$ and $B^{(t)}$ at generation $t$.
Let $\widehat{F}_A^{(t)}(\cdot)$ and $\widehat{F}_B^{(t)}(\cdot)$ denote the (sampling-based) comprehensive fitness used by the governance layer.
Let $m^{(t)}\in B^{(t)}$ denote the current marker opponent and $l\in(0,1)$ the update threshold.

\paragraph{(1) Marker update timing via BufferCount}
At each generation, we compute the 75th-percentile comprehensive fitness value within population $A^{(t)}$, denoted by $\widehat{F}^{(t)}_{A,75}$.
We define \texttt{BufferCount} as the number of \emph{consecutive} generations for which $\widehat{F}^{(t)}_{A,75} > l$ holds.
When \texttt{BufferCount} reaches a preset horizon $T$, MGM triggers a marker selection attempt.

\paragraph{(2) Candidate persistence via KeepCount}
We maintain an integer counter $\texttt{KeepCount}^{(t)}(b)$ for each opponent individual $b\in B^{(t)}$, defined as the number of consecutive generations that $b$ remains in the current opponent population:
\[
\texttt{KeepCount}^{(t+1)}(b)=
\begin{cases}
\texttt{KeepCount}^{(t)}(b)+1, & b\in B^{(t)}\cap B^{(t+1)},\\
1, & b\in B^{(t+1)}\setminus B^{(t)},\\
\text{undefined}, & b\notin B^{(t+1)}.
\end{cases}
\]
When an individual is eliminated, its counter is removed together with the individual.
When selecting a new marker, we restrict candidates to those with $\texttt{KeepCount}^{(t)}(b)\ge T$, ensuring that transient opponents do not cause frequent marker switches.

\paragraph{Parameter coupling with elimination rate}
With per-generation elimination rate $\vartheta$, a natural time horizon is
\begin{equation}
T \;\triangleq\; \left\lceil \frac{1}{\vartheta}\right\rceil,
\end{equation}
which aligns the update timescale with the expected full refresh period of the population under the worst-case replacement scenario.

\paragraph{Selection rule and fallback}
When the update is triggered, MGM selects the new marker from the eligible set by prioritizing persistence and breaking ties by opponent strength:
\[
m^{(t+1)} \in \arg\max_{b\in B^{(t)}:\,\texttt{KeepCount}^{(t)}(b)\ge T}
\Bigl(\texttt{KeepCount}^{(t)}(b),\ \widehat{F}_B^{(t)}(b)\Bigr),
\]
where the maximization is lexicographic.
If no eligible candidate exists, MGM keeps the old marker unchanged.
Algorithm~\ref{alg:marker_gene_update} summarizes the procedure.

\begin{algorithm}[t]
\small
\caption{Marker Gene Update (population $A$ updates marker in $B$)}
\label{alg:marker_gene_update}
\SetAlgoLined
\LinesNumbered

\KwIn{$A^{(t)},B^{(t)}$; marker $m^{(t)}\in B^{(t)}$; threshold $l$; rate $\vartheta$;\;}
\KwIn{\texttt{BufferCount}; counters $\texttt{KeepCount}^{(t)}(\cdot)$.}
\KwOut{$m^{(t+1)}$ and updated \texttt{BufferCount}.}

$T \leftarrow \lceil 1/\vartheta\rceil$\;

\tcp{(1) Fitness evaluation in $A^{(t)}$}
\ForEach{$a\in A^{(t)}$}{
  compute $\widehat{F}_A^{(t)}(a)$\;
}
Sort $A^{(t)}$ by $\widehat{F}_A^{(t)}(a)$ (desc.)\;
Let $\widehat{F}^{(t)}_{A,75}$ be the 75th percentile\;

\tcp{(2) BufferCount (timing gate)}
\If{$\widehat{F}^{(t)}_{A,75} > l$}{
  $\texttt{BufferCount} \leftarrow \texttt{BufferCount} + 1$\;
}\Else{
  $\texttt{BufferCount} \leftarrow 0$\;
}

\tcp{(3) KeepCount (persistence)}
\ForEach{$b\in B^{(t)}$}{
  \eIf{$b \in B^{(t-1)}$}{
    $\texttt{KeepCount}^{(t)}(b) \leftarrow \texttt{KeepCount}^{(t-1)}(b) + 1$\;
  }{
    $\texttt{KeepCount}^{(t)}(b) \leftarrow 1$\;
  }
}

$m^{(t+1)} \leftarrow m^{(t)}$\;

\tcp{(4) Marker update trigger}
\If{$\texttt{BufferCount} \ge T$}{
  $S \leftarrow \{ b\in B^{(t)} \mid \texttt{KeepCount}^{(t)}(b)\ge T \}$\;
  \If{$S \neq \emptyset$}{
    $m^{(t+1)} \leftarrow \texttt{SelectMarker}(S, \texttt{KeepCount}^{(t)}(\cdot))$\;
    $\texttt{BufferCount} \leftarrow 0$\;
  }
}
\Return{$m^{(t+1)}$, \texttt{BufferCount}}\;
\end{algorithm}

\begin{algorithm}[ht]
\caption{Sub-Procedure: \texttt{SelectMarker}}
\label{alg:select_marker}
\SetAlgoLined
\LinesNumbered
\KwIn{Candidate set $S\subseteq B^{(t)}$; counters $\texttt{KeepCount}^{(t)}(\cdot)$.}
\KwOut{Selected marker $b^*\in S$.}

$b^* \leftarrow \text{null}$\;
$\kappa^* \leftarrow -\infty$\;
$f^* \leftarrow -\infty$\;

\ForEach{$b\in S$}{
  $\kappa \leftarrow \texttt{KeepCount}^{(t)}(b)$\;
  retrieve (or compute) $\widehat{F}_B^{(t)}(b)$\;
  \If{$\kappa > \kappa^*$ \textbf{or} ($\kappa = \kappa^*$ \textbf{and} $\widehat{F}_B^{(t)}(b) > f^*$)}{
    $b^* \leftarrow b$\;
    $\kappa^* \leftarrow \kappa$\;
    $f^* \leftarrow \widehat{F}_B^{(t)}(b)$\;
  }
}
\Return{$b^*$}\;
\end{algorithm}

\subsection{Theoretical Stability Analysis}
\label{sec:theory_stability}

To substantiate the effectiveness of the proposed Marker Gene Method (MGM), we analyze its stability and convergence behavior under a tractable game-theoretic setting.
Because rigorous convergence proofs for general competitive co-evolutionary algorithms in arbitrary games are difficult, we focus on a symmetric strictly competitive game (SCG) formulation, which preserves the adversarial structure while enabling precise dynamical analysis.

\paragraph{Evaluation protocol and sampling rate}
Our analysis matches the evaluation procedure used by MGM: in each evaluation round $t$, every individual is evaluated against a randomly resampled subset of opponents drawn from the current opposing population.
For a population size $N_{\mathrm{pop}}$, the number of sampled opponents per individual is set proportional to $N_{\mathrm{pop}}$,
\begin{equation}
\label{eq:k_sampling}
k \;=\; \left\lceil \rho\,N_{\mathrm{pop}} \right\rceil,
\end{equation}
where $\rho\in(0,1]$ is a fixed sampling ratio.
This ``resample-every-round'' design reduces persistent pairing effects and yields payoff observations with weak dependence across matches.
Under the non-transitive interaction structure and random opponent sampling, we can bound evaluation fluctuations using a Hoeffding-type concentration result under weak-dependence conditions (see Supplementary Material).

\paragraph{Main implications}
The theoretical results can be summarized as follows (formal statements and proofs are provided in the Supplementary Material).

\paragraph{(i) Stabilized selection under finite evaluation budgets}
Conditioned on a fixed marker, the marker-based and generalization-based fitness estimates concentrate around their population-level counterparts as $k$ increases.
Consequently, the probability of ranking reversals caused by finite-sample evaluation noise decreases with $k$, which stabilizes the evolutionary direction early in training.

\paragraph{(ii) Conservative marker replacement via windowed updates}
Let $K$ denote the effective marker-update window length (implemented through KeepCount-based persistence thresholds).
The MGM update rule is conservative: it promotes replacing the current marker only when a candidate exhibits both sufficient persistence and superior estimated fitness, making detrimental marker switches increasingly unlikely as $K$ increases and evaluation noise decreases.

\paragraph{(iii) Local exponential stability in the deterministic limit}
Around a Nash equilibrium $x^\ast$, we study the deterministic MGM-induced dynamics on the simplex tangent space.
We show local contraction by proving that the symmetric part of the Jacobian at $x^\ast$ is negative definite, which implies that $x^\ast$ is a locally exponentially stable attractor for the deterministic system.

\paragraph{(iv) Time-scale separation under small stochastic perturbations}
Finally, we model finite-sample evaluation noise as a small stochastic perturbation of the deterministic flow.
Standard small-noise arguments (e.g., Freidlin--Wentzell theory) imply that the typical escape time from a compact neighborhood of $x^\ast$ grows exponentially as the noise level decreases.
As a consequence, the time to enter a compact Nash neighborhood is much shorter than the time to leave a non-Nash region, which is in turn much shorter than the noise-induced escape time from the Nash neighborhood.
This provides a theoretical explanation for the empirical observation that MGM mitigates cycling (e.g., Red Queen dynamics) and improves stability in SCGs.
\subsection{Validation of Theoretical Analysis on SCGs (RPS)}
\label{sec:validation_rps}

\begin{figure}[htbp]
    \centering
    \includegraphics[width=0.9\columnwidth]{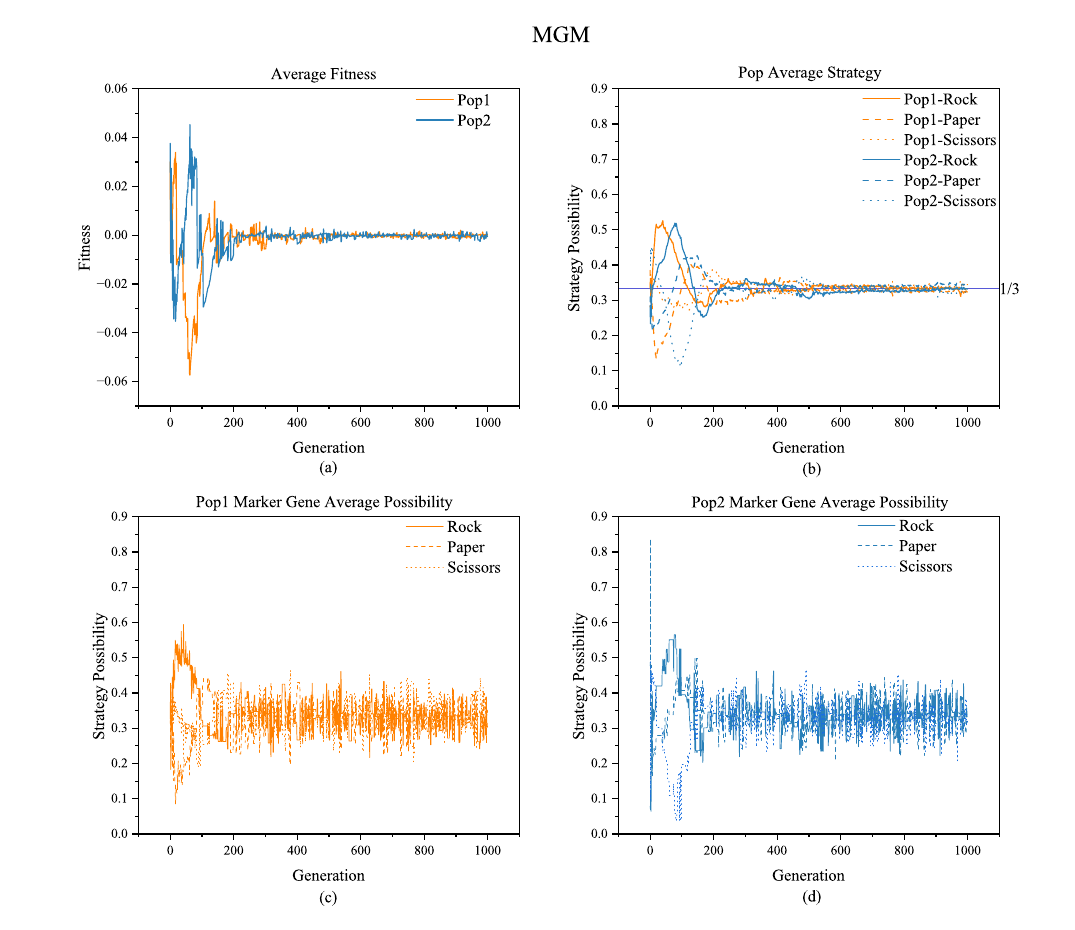}
    \caption{MGM dynamics in RPS over 1000 generations: (a) average fitness for Pop1 and Pop2; (b) average strategy probabilities for Pop1 and Pop2 (dashed line at $1/3$); (c) marker gene strategy probabilities for Pop1; and (d) marker gene strategy probabilities for Pop2.}
    \label{mgm_rps_standard}
\end{figure}
To validate the convergence implications of our theoretical analysis, we conduct evolutionary game experiments on the classic Rock--Paper--Scissors (RPS) model. RPS is a canonical strictly competitive game (SCG) with a unique mixed-strategy Nash equilibrium (NE) and pronounced non-transitivity; under standard competitive coevolutionary algorithms, it commonly induces persistent cyclic dynamics. This makes RPS a suitable benchmark to evaluate whether MGM mitigates cycling and steers the population toward the NE. We also include an ablation on the memory pool (MP) component to assess its contribution; additional details are reported in the Supplementary Material.

\paragraph{Game and representation}
We adopt the standard RPS payoff matrix
\[
\begin{bmatrix}
0 & -1 & 1 \\
1 & 0 & -1 \\
-1 & 1 & 0
\end{bmatrix}.
\]
The unique NE is the mixed strategy that plays each action with probability $1/3$, corresponding (in coevolution) to a population state where the three pure strategies coexist equally.

We use a two-population coevolution model with population sizes $N_1=N_2=100$ unless otherwise stated. Each individual is represented by a 3-dimensional probability vector over $\{\text{Rock},\text{Paper},\text{Scissors}\}$ constrained to be nonnegative and sum to 1. The crossover rate is 1. Mutation is implemented as Gaussian perturbation with strength 0.05 and applied with rate 0.1.

\paragraph{Payoff scale and DWAM target level}
We do not normalize payoffs in the implementation. All reported results use the original payoff $u$. Accordingly, the DWAM target level is set on the same payoff scale and denoted by $l_u$; for RPS we use $l_u=-0.005$, i.e., slightly below the NE payoff, to trigger the intended stability-to-generalization transition. We set the base weight $\omega=0.9$, eliminate the worst 20\% of individuals at each update, and use $KeepCountThreshold=5$ and $BufferCountThreshold=5$.

\paragraph{Baselines and metric}
We compare MGM against three baselines in RPS: Baseline A removes all MGM mechanisms; Baseline B adds a Hall of Fame; Baseline C uses a Pareto-update rule. Convergence is quantified by the KL divergence from the mean population strategy to the NE (Fig.~\ref{comaprison_base}). MGM exhibits a consistent downward trend and stabilizes at a low KL value, while all baselines remain at substantially higher KL values throughout evolution.

Qualitatively, Baselines A and B exhibit persistent intransitive cycles, whereas Baseline C tends to stagnate due to a loss of diversity (supporting visualizations are provided in the Supplementary Material). In contrast, the MGM dynamics (Fig.~\ref{mgm_rps_standard}) converge to a tight neighborhood around the NE. Notably, even after the population has stabilized near the NE, the marker genes (Fig.~\ref{mgm_rps_standard}(c,d)) continue to update frequently with small-amplitude changes. This behavior is consistent with MGM's design: marker updates act as a continuous adversarial ``stress test'' that maintains adaptive pressure, discourages brittle cycling solutions, and favors strategies that generalize across strong opponents, which aligns with the sustained low KL divergence in Fig.~\ref{comaprison_base}.

\paragraph{Population size as a proxy for evaluation intensity}
A key implication of our analysis is that increasing evaluation intensity strengthens stability by reducing disruptive ranking-reversal events; under our evaluation protocol this intensity scales with the number of sampled interactions per individual, $k_{-i}^{(t)}=\lceil \rho N_{-i}^{(t)}\rceil$, and therefore increases with opponent population size. To empirically test this prediction, we repeat the RPS experiment with a reduced population size ($N_1=N_2=50$) while keeping all other settings unchanged, and run 10 independent seeds for each condition.

\begin{figure}[htbp]
    \centering
    \includegraphics[width=0.9\columnwidth]{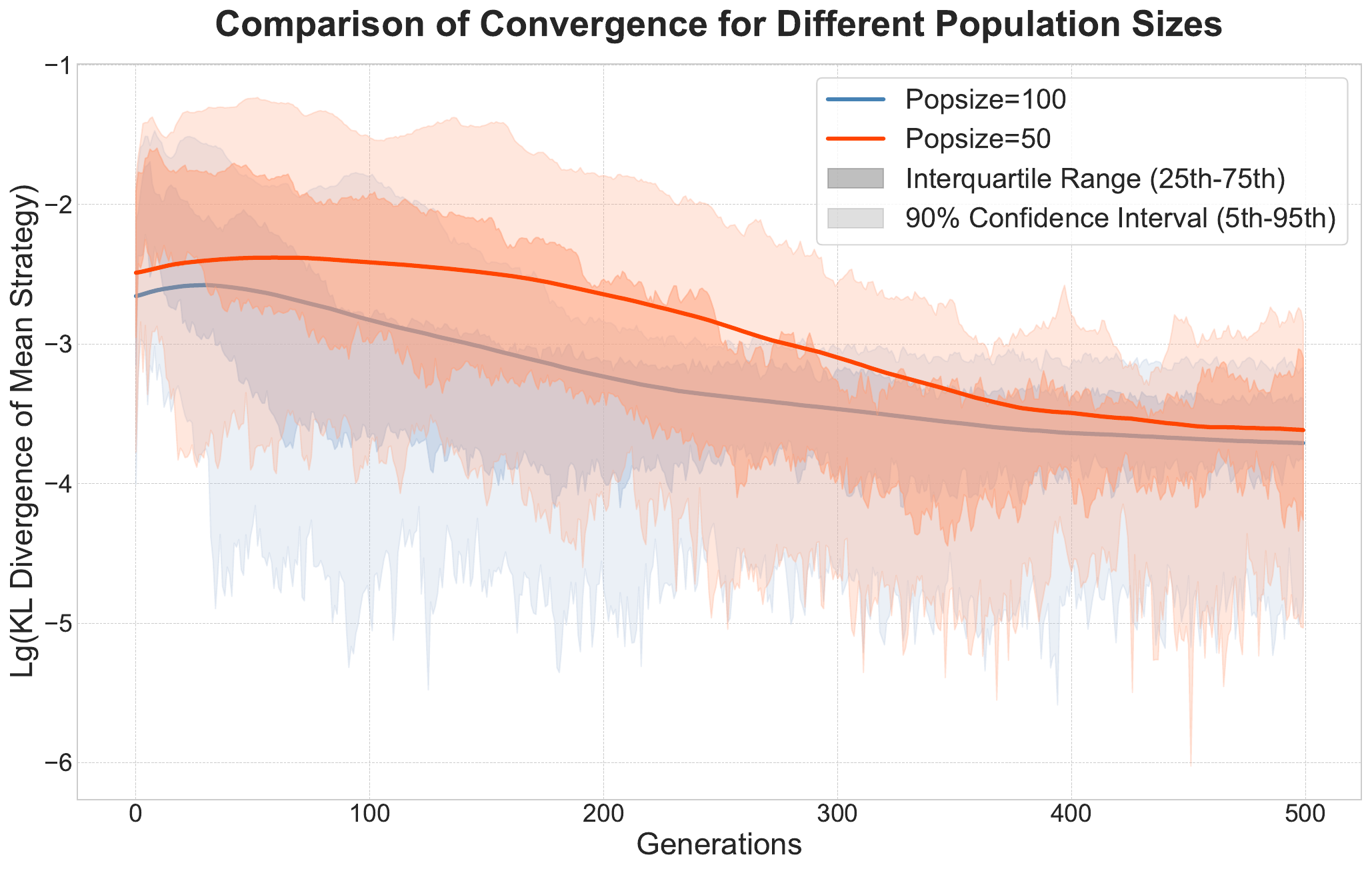}
    \caption{Effect of population size on convergence in the RPS game. The curves show the mean KL divergence from the population's average strategy to the NE over 10 independent runs for $N=100$ and $N=50$. Shaded bands indicate the interquartile range (25th--75th percentiles) and the empirical 5th--95th percentile interval, reflecting stability and run-to-run variability.}
    \label{mgm_rps_popsize_effects}
\end{figure}

As summarized in Fig.~\ref{mgm_rps_popsize_effects}, the larger population ($N=100$) achieves a consistently lower and more stable KL divergence, with narrower percentile bands, whereas the smaller population ($N=50$) shows higher variance and a higher KL level. This supports the theory-aligned prediction that larger populations (hence larger $k_{-i}^{(t)}$ under proportional sampling) provide more reliable fitness estimates and strengthen the effective basin of attraction around the NE. Per-run curves and additional diagnostics are reported in the Supplementary Material.

\subsection{NGD-Div: Divergence-Driven Adaptation of the DWAM Threshold}
\label{sec:ngd_div}

NGD-Div adapts the DWAM threshold $l$ online, aiming to keep the induced co-evolutionary dynamics in a stable regime. Importantly, we do not claim to explicitly compute or directly optimize the true flow divergence $\mathrm{div}\,F$ of the underlying continuous-time dynamics, which is generally intractable in our setting. Instead, NGD-Div minimizes a \emph{computable proxy} whose sign and trend are aligned with the dominant divergence term identified in our local stability analysis (Supplementary Material).

\paragraph{Dynamic equilibrium point (DEP)}
We refer to a \emph{Dynamic Equilibrium Point (DEP)} as a regime where the controller update becomes small while the dynamics are empirically dissipative: (i) the controller gradient is close to zero on average, and (ii) the divergence proxy remains negative (or mildly negative) for sustained iterations. This provides a practical criterion for stable operation without requiring exact divergence estimation.

\paragraph{Controller objective}
We optimize the scalar threshold $l$ by minimizing a smooth control loss
\begin{equation}
\label{eq:control_loss}
    L(l) \;=\; w_A\, A(l)^2 \;+\; w_D\,\big(D(l)-\varepsilon(l)\big)^2 \;+\; w_{\text{anchor}}^{\text{eff}}\,\big(l-l_{\text{anchor}}\big)^2,
\end{equation}
where the three terms respectively (1) encourage DWAM to operate in a desired weighting regime, (2) encourage a dissipative (stable) regime via a divergence-aligned proxy, and (3) stabilize the controller when the gate saturates.

For the first term, let $\alpha_i=\alpha(\beta_i,\sigma_i)$ denote the DWAM selection weight (Eq.~\eqref{eq:dwam_alpha}) for agent $i$ with gate input $\beta_i=\widehat{B}_i-l$. We define
\begin{equation}
\label{eq:A_def}
A(l) \;=\; \alpha_{\mathrm{mean}}(l)-\alpha_{\mathrm{target}},\qquad
\alpha_{\mathrm{mean}}(l)=\frac{1}{N}\sum_{i=1}^N \alpha_i,
\end{equation}
where $\alpha_{\mathrm{target}}$ is a user-chosen target level (e.g., encouraging marker-centric learning early on).

For the anchor term, we use
\begin{equation}
\label{eq:l_anchor}
l_{\text{anchor}} \;=\; \mathbb{E}_{i\in\mathrm{pop}}[\widehat{B}_i],
\end{equation}
which ensures a non-vanishing and smooth gradient signal for $l$ even when some individuals lie in the flat (saturated) region of $\alpha(\cdot)$.

\paragraph{Divergence proxy}
Let $r_i=\widehat{B}_i-\widehat{G}_i$ denote the marker--generalization fitness gap for agent $i$, and $\vec r=(r_1,\ldots,r_N)^\top$.
We define the empirical divergence proxy as
\begin{equation}
\label{eq:empirical_divergence}
    D(l) \;=\; \|\vec r\|_2 \cdot \frac{1}{N}\sum_{i=1}^N \mathrm{sign}(r_i)\,\alpha_i'(l),
\end{equation}
where $\alpha_i'(l)$ denotes the derivative of $\alpha(\beta_i,\sigma_i)$ with respect to its scalar gate input $\beta_i=\widehat{B}_i-l$ (equivalently, with respect to $l$ up to a sign). Intuitively, the averaged signed-slope term measures whether the gate increases selection pressure in a direction that counters destabilizing marker--generalization discrepancies, while $\|\vec r\|_2$ scales the signal by the discrepancy magnitude.

To encourage a mildly dissipative regime while keeping the objective smooth, we use an adaptive negative target
\begin{equation}
\label{eq:eps_adaptive}
\varepsilon(l) \;=\; -\kappa\,|D(l)|,
\end{equation}
with $\kappa>0$, so the controller primarily enforces the \emph{sign preference} ($D<0$) without over-penalizing magnitude.
\begin{figure}
    \centering
    \includegraphics[width=0.9\linewidth]{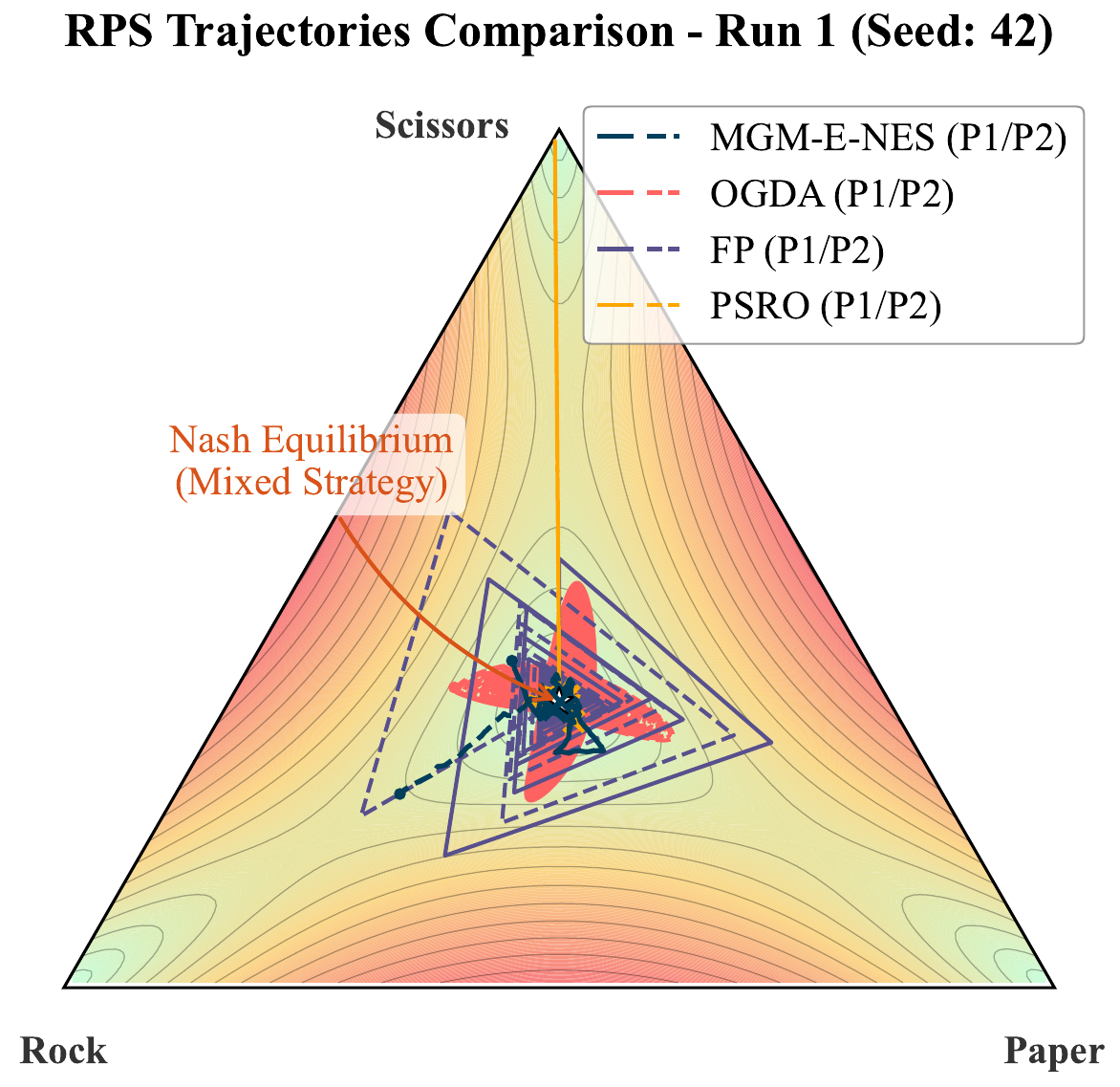}
    \caption{Convergence trajectory against baselines.}
    \label{fig:rps3d_trajectory_single}
\end{figure}
The gradient of $D$ w.r.t.\ $l$ is computed via the curvature of the DWAM gate:
\begin{equation}
\label{eq:divergence_gradient}
    \frac{\partial D}{\partial l}
    \;=\;
    -\|\vec r\|_2 \cdot \frac{1}{N}\sum_{i=1}^N \mathrm{sign}(r_i)\,\alpha_i''(l),
\end{equation}
where $\alpha_i''(l)$ denotes the second derivative of $\alpha(\beta_i,\sigma_i)$ with respect to $\beta_i$.

\paragraph{Preconditioned (natural-gradient style) update}
Finally, we update the controller by a preconditioned gradient step:
\begin{equation}
\label{eq:ngd_div_update}
    l^{(t+1)} \leftarrow l^{(t)} - \eta_l \cdot \mathcal{F}(t)^{-1} \cdot \nabla_{l} L\!\left(l^{(t)}\right),
\end{equation}
where $\eta_l$ is the controller learning rate and $\mathcal{F}(t)>0$ is a scalar uncertainty proxy that damps the update when the population statistics are noisy. In our implementation,
\begin{equation}
\label{eq:fim_proxy}
\mathcal{F}(t)=\mathrm{Var}\!\left[\widehat{B}-\widehat{G}\right] + \tau^2,
\end{equation}
so the effective step size decreases when the marker--generalization gap exhibits high variance. The correspondence to the inertia controller (i.e., how $w_{\text{anchor}}^{\text{eff}}$ is scheduled/adjusted over time) is described in the next subsection.

\paragraph{Inertial regularization (engineering stabilizer)}
\label{ssec:adaptive_lr}
To further enhance the stability of the NGD-Div controller, we add a lightweight engineering module to reduce reactivity to transient disturbances. 
Specifically, we introduce an inertia coefficient $\gamma$ that increases when the controller updates are locally steady and resets when abrupt changes occur. 
It is updated as:
\begin{equation}
\label{eq:gamma_update}
\gamma^{(t+1)} =
\begin{cases}
    \min(\gamma_{\max}, \gamma^{(t)} \cdot k_{\gamma}), & 
        \begin{aligned}[t]
            & \text{if } \left| |\Delta l^{(t)}| - |\Delta l^{(t-1)}| \right| \\
            & \qquad \le \delta_{\text{stable}}
        \end{aligned} \\
    1.0, & \text{otherwise,}
\end{cases}
\end{equation}
where $\Delta l^{(t)} = l^{(t)}-l^{(t-1)}$, $\gamma_{\max}$ caps the inertia, $k_{\gamma}>1$ is the build-up factor, and $\delta_{\text{stable}}$ is a tolerance threshold.

This inertia $\gamma$ modulates the anchor strength in the NGD-Div control loss (Eq.~\eqref{eq:control_loss}) by
\begin{equation}
\label{eq:anchor_eff}
    w_{\text{anchor}}^{\text{eff}} = w_{\text{anchor}}^{\text{base}} \cdot \gamma^{(t+1)}.
\end{equation}

Intuitively, the controller accumulates inertia during stable periods (smoothing $l$ updates), while resetting to $\gamma=1$ to react quickly when persistent instability is detected.
\begin{figure}
    \centering
    \includegraphics[width=0.9\linewidth]{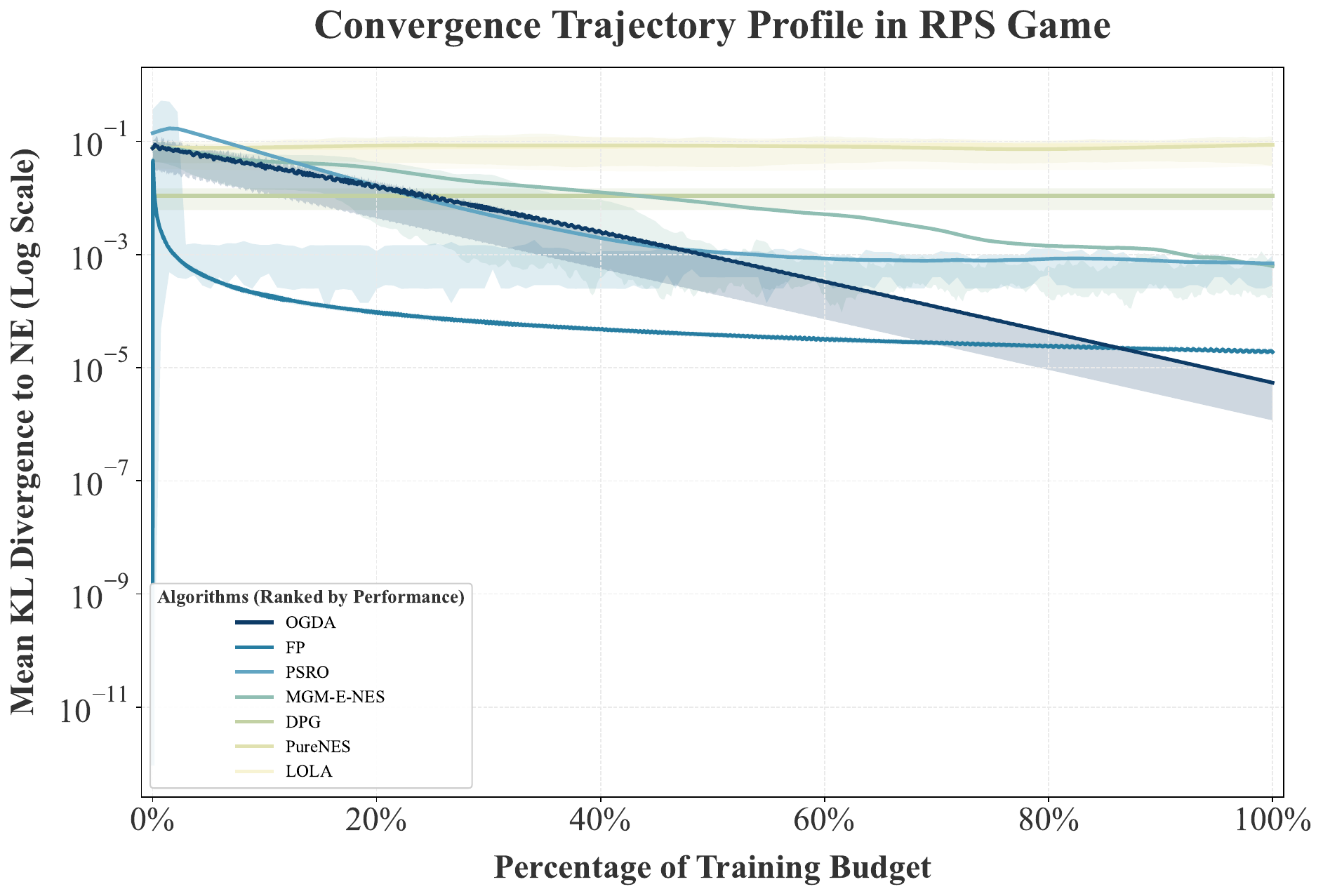}
    \caption{Mean KL divergence to Nash Equilibrium.}
    \label{fig:KLTrajectoryRPS3D}
\end{figure}
\section{The Evolutionary Actuator}


\subsection{The Evolutionary Actuator (NES Instantiation)}
\label{sec:evolutionary_actuator}

Our governance framework (MGM/DWAM/NGD-Div) is agnostic to the underlying optimizer/actuator.
To provide a concrete and reproducible instantiation, we employ a Natural Evolution Strategies (NES) style actuator that updates a search distribution over parameters.
This choice keeps the actuator simple and robust under a fixed rollout budget, while ensuring that our main contributions remain in the governance layer.
\begin{figure}[h!] 
    \centering
    \includegraphics[width=0.9\columnwidth]{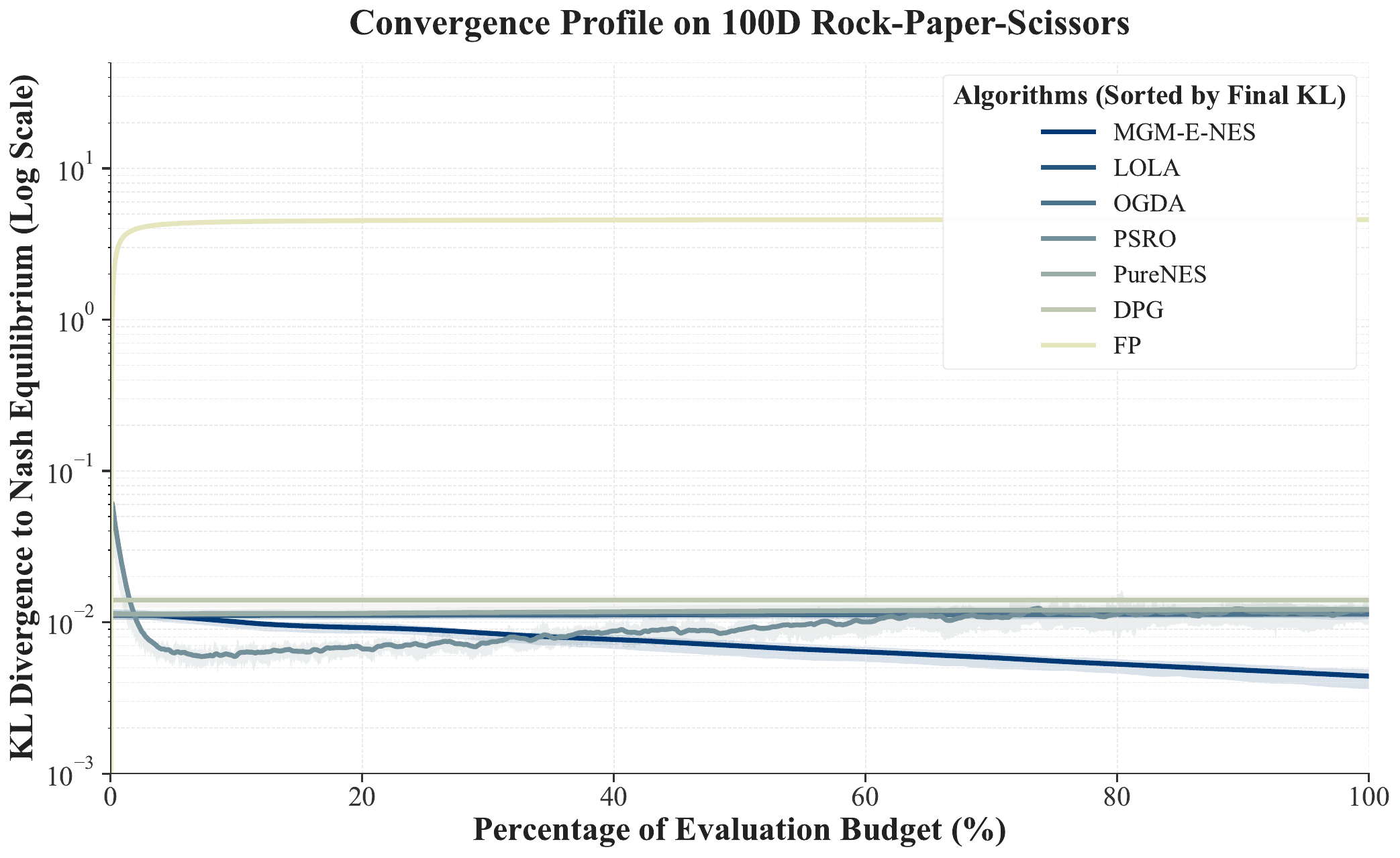}
    \caption{Convergence profile in 100D Rock-Paper-Scissors. Plotted values are the mean KL divergence to the Nash Equilibrium, averaged over both players and multiple runs. (See Supplementary Materials for details).}
    \label{fig:KLTrajectoryRPS100D}
\end{figure}
\paragraph{Search distribution and sampling}
At generation $t$, the actuator maintains a mean parameter vector $\theta^{(t)}\in\mathbb{R}^d$ and an exploration scale $\sigma^{(t)} > 0$, defining an isotropic Gaussian search distribution
$
\theta^{(t)} + \sigma^{(t)}\varepsilon,\ \varepsilon\sim\mathcal{N}(0,I).
$
We draw a population of $N$ perturbations $\{\varepsilon_j\}_{j=1}^{N}$; by default we use antithetic sampling, i.e., sampling $N/2$ i.i.d.\ perturbations and pairing them with their negatives $\{\varepsilon_j,-\varepsilon_j\}$ to reduce gradient-estimation variance.

\paragraph{Fitness signal}

The resulting fitness values $\{f_j^{(t)}\}$ are produced by the governance layer (e.g., via MGM/DWAM-weighted objectives) and treated as black-box rewards by the actuator.

\paragraph{NES gradient estimate and update}
Using the standard score-function estimator for isotropic NES, we compute
\begin{equation}
\widehat{g}^{(t)} \;=\;
\frac{1}{N\,\sigma^{(t)}}\sum_{j=1}^{N} f_j^{(t)}\,\varepsilon_j,
\end{equation}
and update the distribution mean by a fixed learning rate $\eta_\theta$:
\begin{equation}
\theta^{(t+1)} \leftarrow \theta^{(t)} + \eta_\theta \,\widehat{g}^{(t)}.
\end{equation}
When antithetic sampling is enabled, we use the equivalent paired form
$\widehat{g}^{(t)}=\frac{1}{N\,\sigma^{(t)}}\sum_{j=1}^{N/2}(f_j^{(t)}-f_{j+N/2}^{(t)})\,\varepsilon_j$.
Optionally, we apply an extragradient (prediction--correction) step, which empirically improves stability in non-stationary competitive settings.

\paragraph{Why not Adam and classical step-size adaptation (CAS/CSA)}
A natural baseline is to apply Adam-style adaptive learning rates to the NES mean update, and/or to use cumulative step-size adaptation (CSA) to tune $\sigma$.
However, under the same evaluation budget and SCG-induced non-stationarity, these off-the-shelf components can significantly increase run-to-run variance and occasionally destabilize training in our setting (see ablations in Supplementary Material).
Therefore, our default MGM-NES implementation \emph{does not} use Adam and \emph{does not} use CSA.
\begin{figure}
    \centering
        \includegraphics[width=\linewidth]{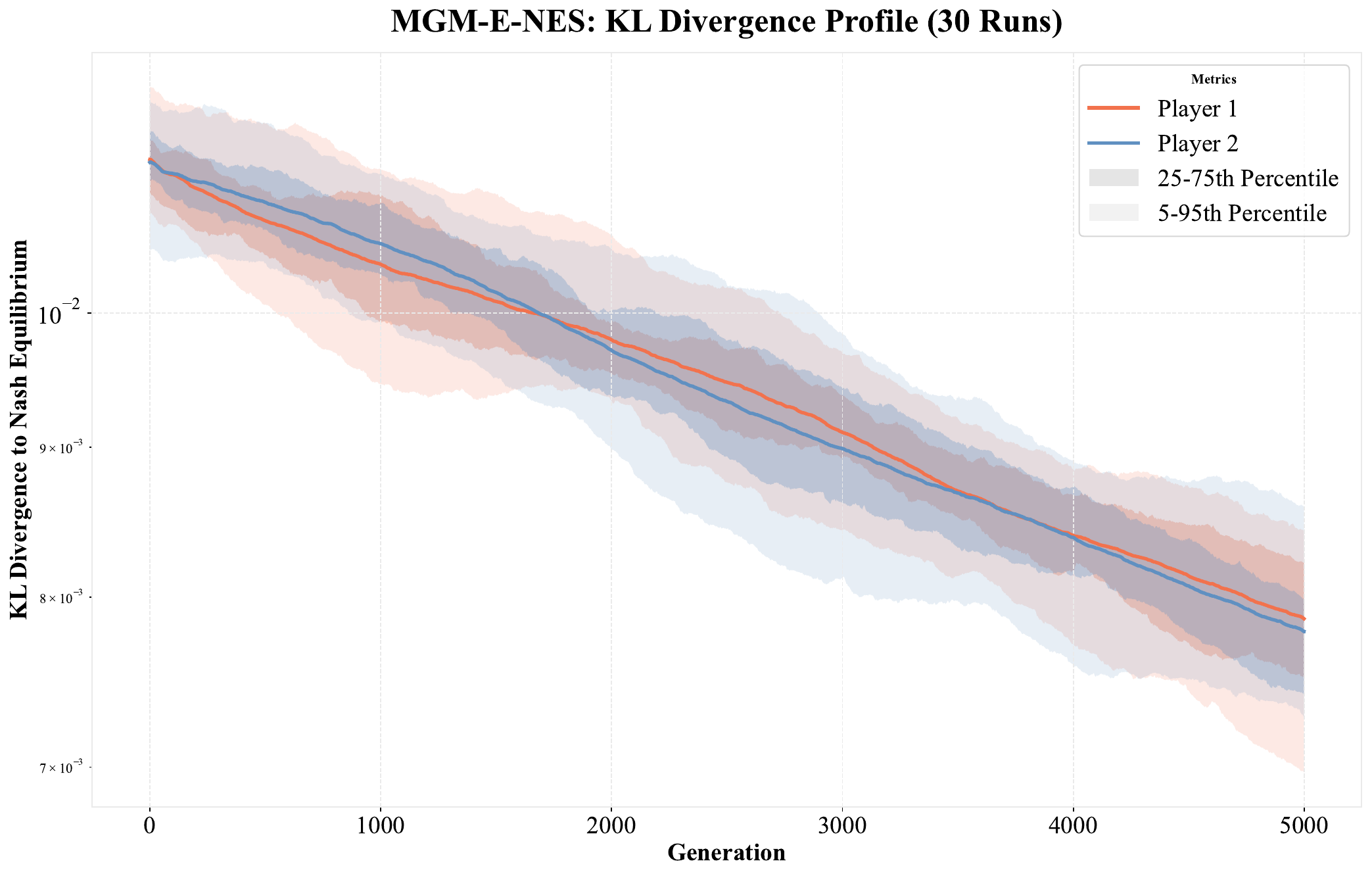}
        \caption{MGM-E-NES KL Divergence Profile (30 Runs) in 1000-Dimension RPS Game.}
        \label{fig:kl_RPS1000}
\end{figure}
\paragraph{Adaptive Exploration Controller (AEC; progress-gated entropy-driven adaptation)}
To improve robustness under a fixed rollout budget, our MGM-NES implementation uses a lightweight controller to adapt the exploration scale $\sigma$ while keeping the mean learning rate $\eta_\theta$ fixed (i.e., no adaptive learning-rate optimizer such as Adam).
Let $\mu_f^{(t)}$ denote the mean governed fitness of the sampled candidates at generation $t$.
We maintain an exponential moving average (EMA) baseline
\begin{equation}
\bar{f}^{(t)} \leftarrow (1-\alpha_{\mathrm{ema}})\bar{f}^{(t-1)} + \alpha_{\mathrm{ema}}\,\mu_f^{(t)},
\end{equation}
and define a progress gate $\mathbb{I}[\mu_f^{(t)} > \bar{f}^{(t)}]$.

If progress is made, AEC anneals exploration by setting the target $\sigma_{\mathrm{tar}} \leftarrow \sigma_{\min}$.
Otherwise, it uses the entropy of the current parameterized policy to decide the exploration intensity.
Specifically, we compute the Shannon entropy of the current policy vector $p$ (obtained from the current parameters, clipped and normalized):
$
H(p)=-\sum_k p_k\log p_k,
$
and normalize it by the maximum entropy $\log(d)$ with $d$ the parameter dimension:
$
\widetilde{H}=H(p)/\log(d).
$
If $\widetilde{H}<0.5$ (low-entropy stagnation), we set $\sigma_{\mathrm{tar}}\leftarrow\sigma_{\max}$; otherwise, $\sigma_{\mathrm{tar}}\leftarrow\sigma_{\mathrm{mid}}$.

Finally, $\sigma$ is updated smoothly toward the target:
\begin{equation}
\sigma^{(t+1)} \leftarrow 0.9\,\sigma^{(t)} + 0.1\,\sigma_{\mathrm{tar}}.
\end{equation}
All implementation details follow the released code.

\subsection{The MGM-E-NES Architecture}
\label{ssec:mgm-e-nes-arch}

Adapting the original MGM framework to a Natural Evolution Strategies (NES) actuator introduces a practical mismatch.
MGM relies on cross-generational anchoring through a marker gene, whereas NES typically performs a \emph{hard update}, sampling a fresh batch of candidates from an updated search distribution each generation.
Without additional design, the reference marker would drift too rapidly to serve as a stable anchor for governance.

\paragraph{Archive-based anchoring}
We resolve this mismatch by introducing an \emph{archive} mechanism, inspired by the Hall-of-Fame idea in coevolutionary learning~\cite{Zychowski2024Coevolutionary}.
At each generation, we identify the candidate with the highest \emph{generalization fitness} (defined in Sec.~\ref{sec:III}) and insert it into an archive $\mathcal{A}$.\footnote{In our implementation, $\mathcal{A}$ is maintained as a bounded buffer of size $H$ (e.g., FIFO or top-$H$ by generalization fitness) to keep memory usage constant.}
When the governance logic requests a new marker gene, we sample a marker candidate uniformly at random from $\mathcal{A}$.
This design decouples marker stability from the rapid turnover of the NES population, providing a relatively stationary evaluation reference while preserving the exploration benefits of distribution-based search.

Algorithm~\ref{alg:mgm_e_nes_A}--\ref{alg:mgm_e_nes_B} summarize the complete MGM-E-NES procedure for player~$i$ (the opponent is symmetric), including DWAM-based evaluation, archive maintenance, archive-based marker rollback, and the NGD-Div update of the threshold parameter~$l$.

\paragraph{Computational complexity and practical cost}
Let $N$ be the NES population size, and let $\rho\in(0,1]$ be the opponent sampling ratio in the evaluation protocol.
For an individual evaluated against the opponent pool of size $N_{-i}$, we sample
$k=\lceil \rho N_{-i}\rceil$ opponents uniformly without replacement.
Let $T$ denote the cost of one rollout (one game episode), and let $D$ be the policy parameter dimension.

Per generation, MGM-E-NES evaluates $N$ sampled candidates against $k$ opponents each, and (when needed for anchoring) additionally evaluates the marker gene against another $k$ opponents.
Therefore, the rollout count per generation scales as
\[
\Theta(Nk + k)=\Theta\big((N+1)\,k\big)=\Theta\big((N+1)\,\rho\,N_{-i}\big),
\]
which becomes $\mathcal{O}(\rho N^2)$ when $N_{-i}=\Theta(N)$; rollout evaluation dominates wall-clock time and is embarrassingly parallel.

The NES distribution update (e.g., weighted aggregation of perturbations) costs $\mathcal{O}(ND)$ per generation.
The governance layer (DWAM and NGD-Div) adds only lightweight bookkeeping plus constant-time arithmetic on a single scalar threshold $l$.
\begin{algorithm}[t]
\small
\caption{MGM-E-NES with archive-based marker rollback (player $i$): evaluation and policy update}
\label{alg:mgm_e_nes_A}
\SetAlgoLined
\LinesNumbered
\DontPrintSemicolon
\setlength{\algomargin}{0.8em}
\SetInd{0.6em}{0.8em}

\KwIn{Population size $N$; opponent size $N_{-i}$; sampling ratio $\rho$; archive cap. $H$ (FIFO); threshold $l^{(t)}$.}
\KwIn{DWAM; NES/AEC hyperparameters.}
\KwOut{$\psi^{(t+1)}$, updated archive $\mathcal{A}$.}

Sample $\theta^{(t)}_{1:N} \sim \pi_{\psi^{(t)}}$\;
$k \leftarrow \lceil \rho N_{-i}\rceil$\;

\For(\tcp*[f]{evaluate}){$j \leftarrow 1$ \KwTo $N$}{
  Sample opponents $\mathcal{O}_j^{(t)}$ of size $k$\;
  $\hat B_j^{(t)} \leftarrow \textsc{EvalBase}\!\left(\theta_j^{(t)}, M^{(t)}\right)$\;
  $\hat G_j^{(t)} \leftarrow \textsc{EvalGen}\!\left(\theta_j^{(t)}, \mathcal{O}_j^{(t)}\right)$\;
  $\alpha_j^{(t)} \leftarrow \alpha\!\left(\hat B_j^{(t)}-l^{(t)},\, \hat G_j^{(t)}-l^{(t)}\right)$\;
  $\hat F_j^{(t)} \leftarrow \alpha_j^{(t)}\hat B_j^{(t)} + (1-\alpha_j^{(t)})\hat G_j^{(t)}$\;
}

$\psi^{(t+1)} \leftarrow \textsc{NESUpdate}\!\left(\psi^{(t)}; \{\theta_j^{(t)},\hat F_j^{(t)}\}_{j=1}^N\right)$\;

$\bar F^{(t)} \leftarrow \frac{1}{N}\sum_{j=1}^N \hat F_j^{(t)}$\;
$\sigma^{(t+1)} \leftarrow \textsc{AECUpdateSigma}\!\left(\sigma^{(t)}, \bar F^{(t)}\right)$\;

$\mathcal{C} \leftarrow \{j:\hat F_j^{(t)} > l^{(t)}\}$\;
\If{$\mathcal{C}\neq \emptyset$}{
  $j^\star \leftarrow \arg\max_{j\in\mathcal{C}} \hat G_j^{(t)}$\;
  Append $\theta_{j^\star}^{(t)}$ to $\mathcal{A}$\;
  \If{$|\mathcal{A}|>H$}{drop the oldest element\;}
}
\end{algorithm}

\begin{algorithm}[t]
\small
\caption{MGM-E-NES governance (player $i$): marker rollback and NGD-Div threshold update}
\label{alg:mgm_e_nes_B}
\SetAlgoLined
\LinesNumbered
\DontPrintSemicolon
\setlength{\algomargin}{0.8em}
\SetInd{0.6em}{0.8em}

\KwIn{Marker $M^{(t)}$; archive $\mathcal{A}$; trigger counter $c$; horizon $K$; threshold $l^{(t)}$.}
\KwOut{$M^{(t+1)}$, $l^{(t+1)}$, updated $c$.}

Sample opponents $\mathcal{O}_M^{(t)}$ of size $k$\;
$\hat G_M^{(t)} \leftarrow \textsc{EvalGen}\!\left(M^{(t)}, \mathcal{O}_M^{(t)}\right)$\;

$F_{\max}^{(t)} \leftarrow \max_j \hat F_j^{(t)}$\;
\If{$F_{\max}^{(t)} > l^{(t)}$}{$c \leftarrow c+1$\;}
\Else{$c \leftarrow 0$\;}

\If(\tcp*[f]{rollback}){$c \ge K \ \textbf{and} \ |\mathcal{A}|>0$}{
  $M^{(t+1)} \leftarrow \textsc{Sample}(\mathcal{A})$\;
  $c \leftarrow 0$\;
}
\Else{
  $M^{(t+1)} \leftarrow M^{(t)}$\;
}

$l^{(t+1)} \leftarrow l^{(t)} - \eta_l \cdot (\mathcal{F}^{(t)})^{-1}\nabla_l L\!\left(l^{(t)}\right)$\;
\end{algorithm}

\section{Case Study}
\label{sec:case_study}

\begin{figure}
        \centering
        \includegraphics[width=0.9\linewidth]{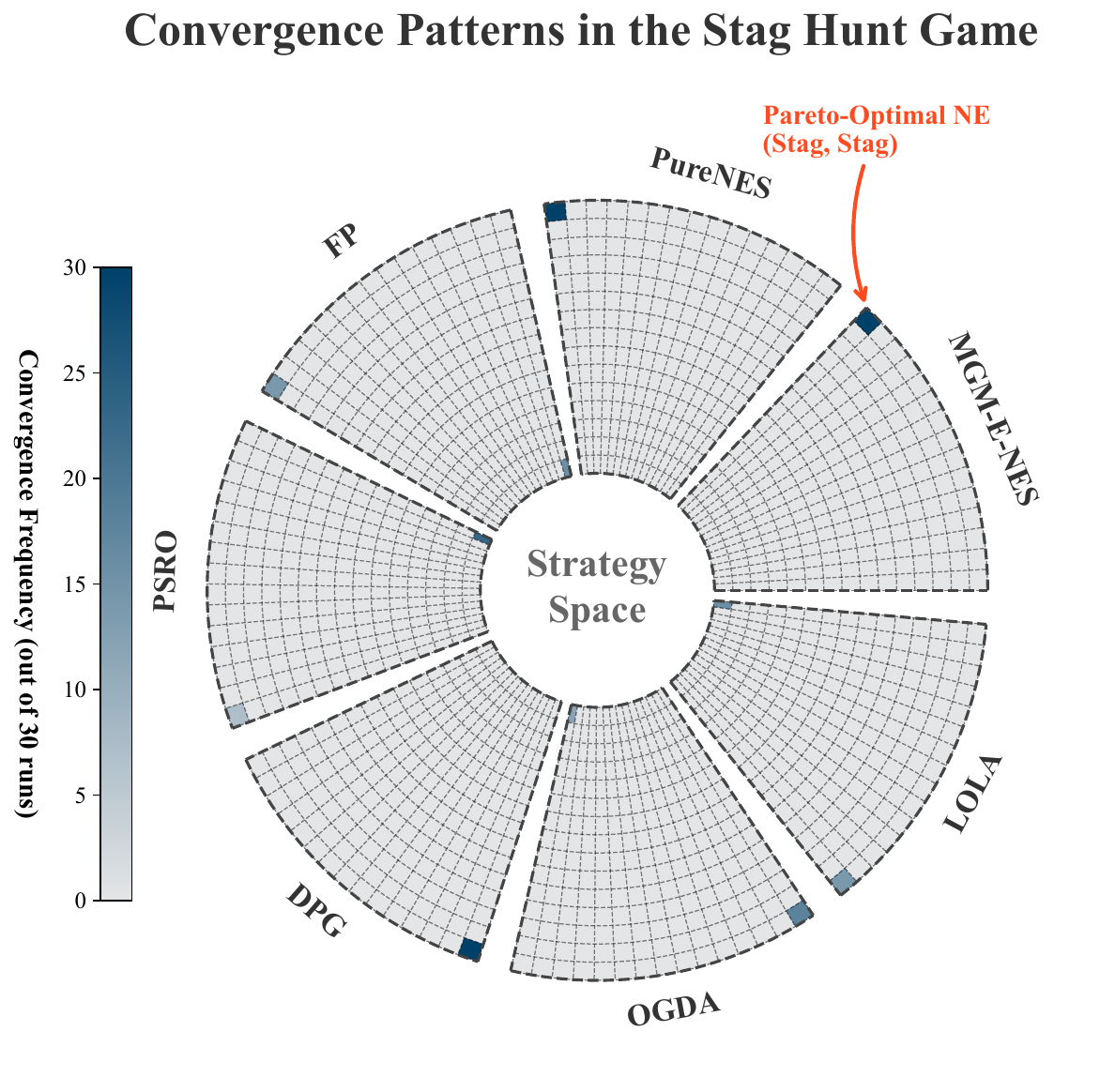}
        \caption{Convergence patterns in the Stag Hunt Game. The heatmap shows the final strategy distribution over 30 runs.}
        \label{fig:stag_hotplot}
\end{figure}

To comprehensively evaluate MGM-E-NES, we benchmarked it against a suite of representative algorithms: Fictitious Play (FP), PSRO , LOLA , Deep Policy Gradient (DPG) and OGDA . 

\paragraph{Ablation Study via PureNES} To isolate and quantify the benefits conferred by our MGM-E control layer, we introduce a critical baseline: PureNES. This algorithm is identical to the NES actuator used in MGM-E-NES, including its adaptive $\sigma$ mechanism, but with the top-level MGM-E controller entirely removed. PureNES thus serves as a direct ablation study of our core contribution.

\paragraph{Experimental Protocol} Our evaluation spans two major game classes: intransitive zero-sum games (3D, 100D, and 1000D Rock-Paper-Scissors) and coordination games (Stag Hunt, Battle of the Sexes). For rigor, each experiment was conducted for 30 independent runs with distinct random seeds. A uniform evaluation budget was allocated to all algorithms for each game. 

\paragraph{Initialization Protocol} To ensure a fair and reproducible comparison, we adopted a differentiated initialization strategy tailored to the nature of each algorithm:
\begin{itemize}
    \item For methods with explicit policy parameterization (LOLA, OGDA, MGM-E-NES, PureNES), we initiated them from the identical random policy parameters for each run, drawn from a normal distribution $\mathcal{N}(0, \sigma^2)$. We set $\sigma=0.5$ for 3D-RPS to provide a clear visualization from a distant starting point, and $\sigma=0.15$ for all other games.
    \item For history-based methods (FP, PSRO), we followed their standard canonical protocols as established in the literature. For neural network-based baselines not covered above, we employed the standard Kaiming initialization.
\end{itemize}

For computational tractability in high-dimensional settings, a sparsified payoff matrix was used. The MGM-E controller's parameters remained fixed across all tasks; only the underlying NES learning rate was conventionally tuned per game. Detailed hyperparameters and pseudocode are provided in the Supplemental Material.

\subsection{Equilibrium Discovery in Intransitive Zero-Sum Games}

\subsubsection{3-Dimension RPS Game}
Figure~\ref{fig:rps3d_trajectory_single} displays a representative trajectory from a convergent run in the 3D Rock-Paper-Scissors (RPS) game, specifically using the first random seed to prevent selection bias. For comprehensive validation, the complete results from all 30 independent runs, including raw data and analysis scripts, are provided in our supplementary material.

The figure shows that MGM-E-NES converges towards the Nash Equilibrium (NE) with a stable and direct trajectory. In contrast, gradient-based methods such as OGDA and FP exhibit characteristic spiraling dynamics, with FP's best-response mechanism leading to a more angular path. PSRO also converges effectively, though its trajectory displays some exploratory deviation. The policies of LOLA are maintained within a neighborhood closer to the NE than those of PureNES.

In terms of convergence speed, OGDA and PSRO outperform other methods in this low-dimensional environment. The stability of the convergence path demonstrated by MGM-E-NES is, however, noteworthy. These differing dynamic behaviors provide context for evaluating algorithm performance in the more challenging, higher-dimensional games that follow.

Detailed convergence results and heatmap visualizations for all 30 independent runs of MGM-E-NES are presented in the Supplemental Material.

The KL divergence trends (Figure~\ref{fig:KLTrajectoryRPS3D}) show that FP, PSRO, and OGDA also converged, with FP achieving the tightest convergence. LOLA maintained its policies in a neighborhood closer to the NE than PureNES. In terms of convergence efficiency, FP performed best, followed by PSRO and MGM-E-NES.

\subsubsection{100-Dimension RPS Game}
To further test the robustness of MGM-E-NES in high-dimensional intransitive games, we increased the dimensionality to 100. In this setting, visualizing policy trajectories is no longer meaningful, so we focus on comparing KL divergence.

As depicted in Figure~\ref{fig:KLTrajectoryRPS100D}, only MGM-E-NES  exhibited a clear and consistent convergence trend. Algorithms that performed well in the 3D environment (e.g., FP,  OGDA) struggled in this high-dimensional challenge and failed to converge effectively . 

\subsubsection{1000-Dimension RPS Game}

Table~\ref{tab:ultimate_judgement} details the performance of all algorithms in the 1000-dimensional RPS game. In this high-dimensional and complex environment, oracle-based methods like PSRO demonstrated a clear advantage in converging towards the Nash Equilibrium, a result consistent with their theoretical design. 
In contrast, MGM-E-NES exhibited a performance ceiling, suggesting that for extremely high-dimensional, purely competitive scenarios, its sample efficiency might be lower than specialized gradient-solving algorithms. Future work could potentially integrate paradigms like Evolutionary Reinforcement Learning to enhance convergence. Figure~\ref{fig:kl_RPS1000} further illustrates the gradual downward trend of KL divergence for MGM-E-NES, confirming its inherent stability even in this highly challenging environment.

\subsubsection{Conclusion on RPS Games}

The series of experiments across 3D, 100D, and 1000D RPS provides a systematic evaluation of MGM-E-NES's performance in intransitive zero-sum games of increasing complexity. The framework consistently exhibits a stable convergence trend across all tested dimensionalities, effectively suppressing policy cycling. Notably, this performance is achieved without any task-specific adjustments to the core MGM-E controller parameters, highlighting its adaptive design. Furthermore, as shown in Table~\ref{tab:ultimate_judgement}, the target value $l$ consistently converges to a value  that is slightly negative but statistically indistinguishable from zero , which demonstrates the NGD-Div optimizer's capability to autonomously identify the correct divergence objective corresponding to the Nash Equilibrium.
\begin{figure}
        \centering
        \includegraphics[width=0.9\linewidth]{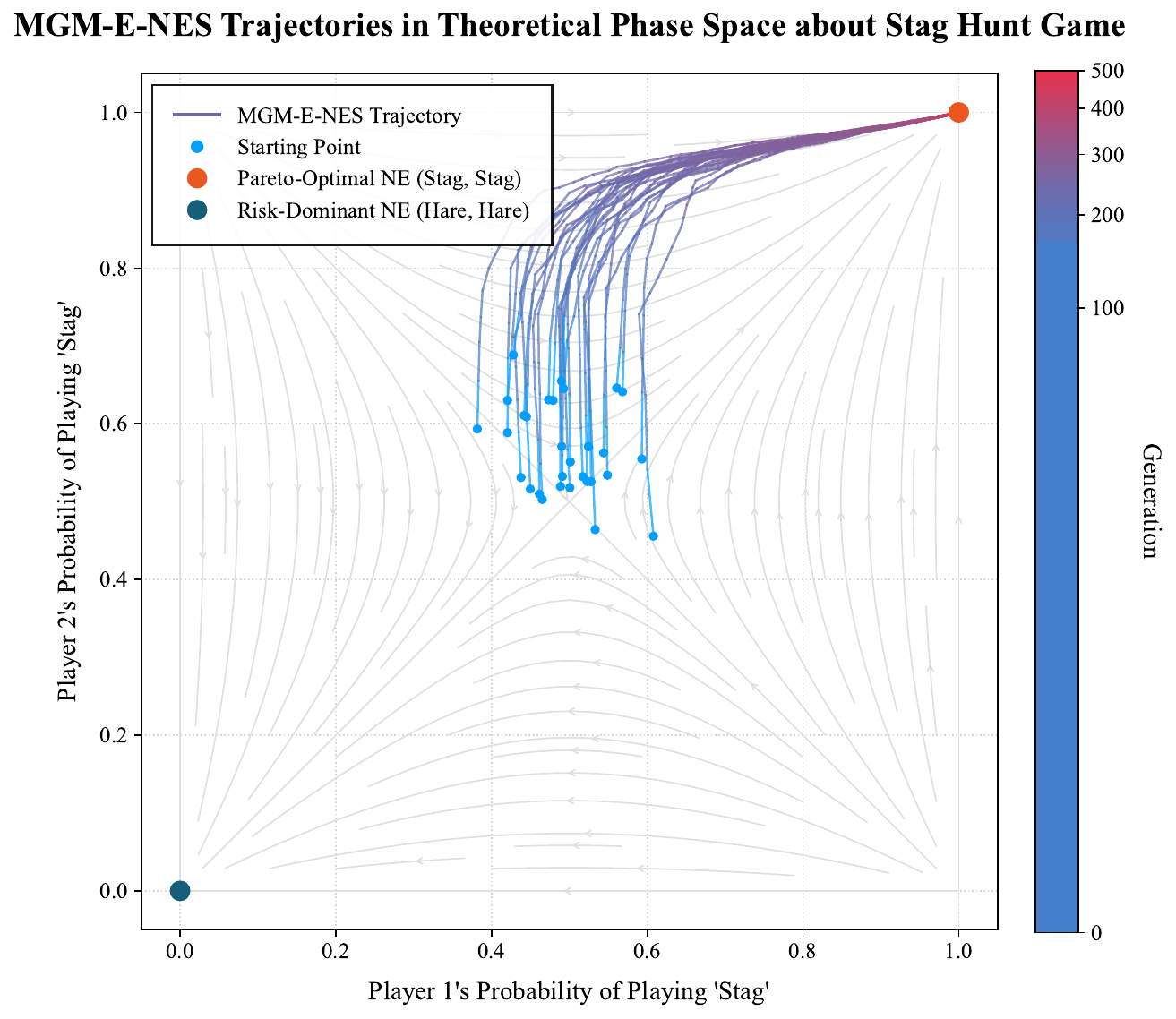}
        \caption{MGM-E-NES trajectories in the strategy simplex for Stag Hunt, converging to the Pareto-optimal NE.}
        \label{fig:stag_hunt_simplex}
\end{figure}
While certain baseline algorithms, such as FP and OGDA, show strong performance in the low-dimensional setting, their efficacy tends to degrade as dimensionality increases. We hypothesize this is attributable to the policy phase space becoming increasingly complex and sensitive to parameter adjustments. Consequently, algorithms that rely on an inward-spiraling convergence dynamic may find their trajectories degenerating into prolonged drift within this vast space, leading to convergence failure or requiring an exceptionally large number of steps.

In contrast, methods like MGM-E-NES and PSRO, which appear to follow more goal-directed convergence paths, maintain their performance more effectively in these challenging high-dimensional landscapes. Empirically, we also note that increasing the population size can further enhance the convergence of MGM-E-NES, with detailed results provided in the Supplemental Material.

\subsection{Equilibrium Discovery in Coordination Games}

The Stag Hunt game is a another classic example of a coordination game, where the core challenge lies in the trade-off between risk and reward. We employed a payoff matrix with high risk and high reward (payoff is 5 for mutual stag hunting; if one hunts stag and the other hare, the stag hunter gets 0 and the hare hunter gets 3; mutual hare hunting yields 2).

In this setting, MGM-E-NES, PureNES, and DPG all stably converged to the Pareto-optimal "Stag Hunt" equilibrium with a success rate of \textbf{$1.00 \pm 0.00$} over 30 independent runs (convergence criterion: $p(\text{Stag}) \ge 0.9$ for both players). The other algorithms failed to achieve consistent convergence. Notably, while DPG succeeded here, it failed to replicate this success in the Battle of the Sexes game, leading us to infer that the specific structure of this Stag Hunt instance may have been particularly favorable for its convergence.

The convergence heatmap in Figure~\ref{fig:stag_hotplot} and the strategy space trajectories in Figure~\ref{fig:stag_hunt_simplex} jointly demonstrate the strong attractive force of MGM-E-NES. Starting from random initializations, its trajectories are powerfully drawn towards the Pareto-optimal Nash Equilibrium, without any policy jumps or hesitation. MGM-E-NES exhibited similar convergence characteristics in the Battle of the Sexes game, which are not detailed here for brevity but are available in the Supplemental Material.

\subsection{Application to a Markov Resource Game}
\label{sec:markov_game}

\subsubsection{Game overview}
We evaluate in a symmetric two-player Markov game that captures resource-dependent incentives and non-stationary evaluation.
The environment has three resource states
\(\mathcal{S}=\{\textsc{Rich},\textsc{Poor},\textsc{Collapsed}\}\).
At each round, both players simultaneously choose an action from
\(\mathcal{A}=\{C,D\}\) (cooperate/defect), obtain state-dependent payoffs, and the state transitions according to the joint action distribution induced by their policies.

Each agent uses a stationary stochastic policy parameterized by a 3-dimensional vector
\[
p=\big(p_R,p_P,p_D\big)\in[0,1]^3,
\qquad
p_s := \Pr(C\mid s),
\]
with \(s\in\{R,P,D\}\) indexing \textsc{Rich}, \textsc{Poor}, and \textsc{Collapsed}.
The opponent is parameterized analogously by \(q\).
In state \(s\), the induced mixed action distributions are
\(\pi_p(s)=[p_s,1-p_s]\) and \(\pi_q(s)=[q_s,1-q_s]\) over \([C,D]\).

To avoid degenerate deterministic transitions and to model evaluation noise, we apply a trembling-hand perturbation with rate \(\epsilon\):
\[
p \leftarrow (1-\epsilon)p + \tfrac{\epsilon}{2}\mathbf{1},\qquad
q \leftarrow (1-\epsilon)q + \tfrac{\epsilon}{2}\mathbf{1},
\]
with \(\epsilon=0.01\) unless stated otherwise.

The row player’s one-step payoff in state \(s\) is given by a \(2\times 2\) matrix \(A_s\)
(rows/columns ordered as \(\{C,D\}\)):
\[
A_{\textsc{Rich}}=
\begin{bmatrix}
4.0 & 0.0\\
5.0 & 1.0
\end{bmatrix},\quad
A_{\textsc{Poor}}=
\begin{bmatrix}
2.0 & 0.0\\
3.0 & 0.5
\end{bmatrix},\quad
A_{\textsc{Collapsed}}=
\begin{bmatrix}
0.5 & -0.5\\
1.0 & 0.0
\end{bmatrix}.
\]
We assume a symmetric game and define the column player’s payoff matrix as \(A_s^\top\).
Thus, for a fixed state \(s\), the expected stage payoffs are
\[
r_1(s;p,q) = \pi_p(s)^\top A_s \,\pi_q(s),\qquad
r_2(s;p,q) = \pi_q(s)^\top A_s^\top \,\pi_p(s).
\]

Let \(M(p,q)\in\mathbb{R}^{3\times 3}\) denote the state transition matrix induced by \((p,q)\),
with indexing \(0=\textsc{Rich},\,1=\textsc{Poor},\,2=\textsc{Collapsed}\) and
\(
M_{ij}(p,q):=\Pr(s_{t+1}=j\mid s_t=i,p,q).
\)
The transition rules are:
\begin{itemize}
    \item \textbf{From \textsc{Rich} (state 0):}
    let \(\Pr(DD\mid \textsc{Rich})=(1-p_R)(1-q_R)\). A double defection downgrades the state to \textsc{Poor}, otherwise the system remains \textsc{Rich}:
    \[
    M_{0,1} = (1-p_R)(1-q_R),\qquad
    M_{0,0} = 1 - M_{0,1},\qquad
    M_{0,2}=0.
    \]
    \item \textbf{From \textsc{Poor} (state 1):}
    let \(\Pr(CC\mid \textsc{Poor})=p_P q_P\) and \(\Pr(DD\mid \textsc{Poor})=(1-p_P)(1-q_P)\).
    Double cooperation restores \textsc{Rich} with probability \(0.8\); double defection collapses to \textsc{Collapsed} with probability \(1\); all other outcomes keep the state \textsc{Poor}:
    \[
    M_{1,0} = 0.8\,p_P q_P,\qquad
    M_{1,2} = (1-p_P)(1-q_P),\qquad
    M_{1,1} = 1 - M_{1,0} - M_{1,2}.
    \]
    \item \textbf{From \textsc{Collapsed} (state 2):}
    let \(\Pr(CC\mid \textsc{Collapsed})=p_D q_D\). Only double cooperation can recover, moving to \textsc{Poor} with probability \(0.2\); otherwise the system remains \textsc{Collapsed}:
    \[
    M_{2,1} = 0.2\,p_D q_D,\qquad
    M_{2,2} = 1 - M_{2,1},\qquad
    M_{2,0}=0.
    \]
\end{itemize}

For a fixed policy pair \((p,q)\), we compute an approximate stationary distribution \(\mu(p,q)\) by power iteration starting from the uniform distribution:
\[
\mu^{(0)} = (1/3,1/3,1/3),\qquad
\mu^{(t+1)} = \mu^{(t)} M(p,q),\qquad t=0,\dots,49.
\]
We then score each player by the stationary-weighted expected stage payoff:
\[
J_1(p,q)=\sum_{s\in\mathcal{S}} \mu_s(p,q)\, r_1(s;p,q),\qquad
J_2(p,q)=\sum_{s\in\mathcal{S}} \mu_s(p,q)\, r_2(s;p,q).
\]

This environment couples short-term incentives to long-term consequences through the endogenous state distribution \(\mu(p,q)\).
As the opponent population drifts, the induced transition matrix \(M(p,q)\) and visitation frequencies over \textsc{Rich}/\textsc{Poor}/\textsc{Collapsed} change, which can substantially alter measured fitness and thus destabilize selection under noisy rollouts.
This makes the resource Markov game a natural testbed for black-box coevolutionary methods that aim to stabilize progress signals across generations.

\subsubsection{Results}

Importantly, this Markov resource game uses the same low-dimensional policy parameterization as our coordination-game and 3D-RPS experiments. We therefore keep the entire MGM-E-NES hyperparameter configuration identical to that used in the coordination game, with no task-specific retuning. This serves as a robustness check: any gains in this Markov game are attributable to MGM’s cross-generational control mechanism rather than per-environment hyperparameter search.
\begin{figure}[h]
    \centering
    \includegraphics[width=1.0\linewidth]{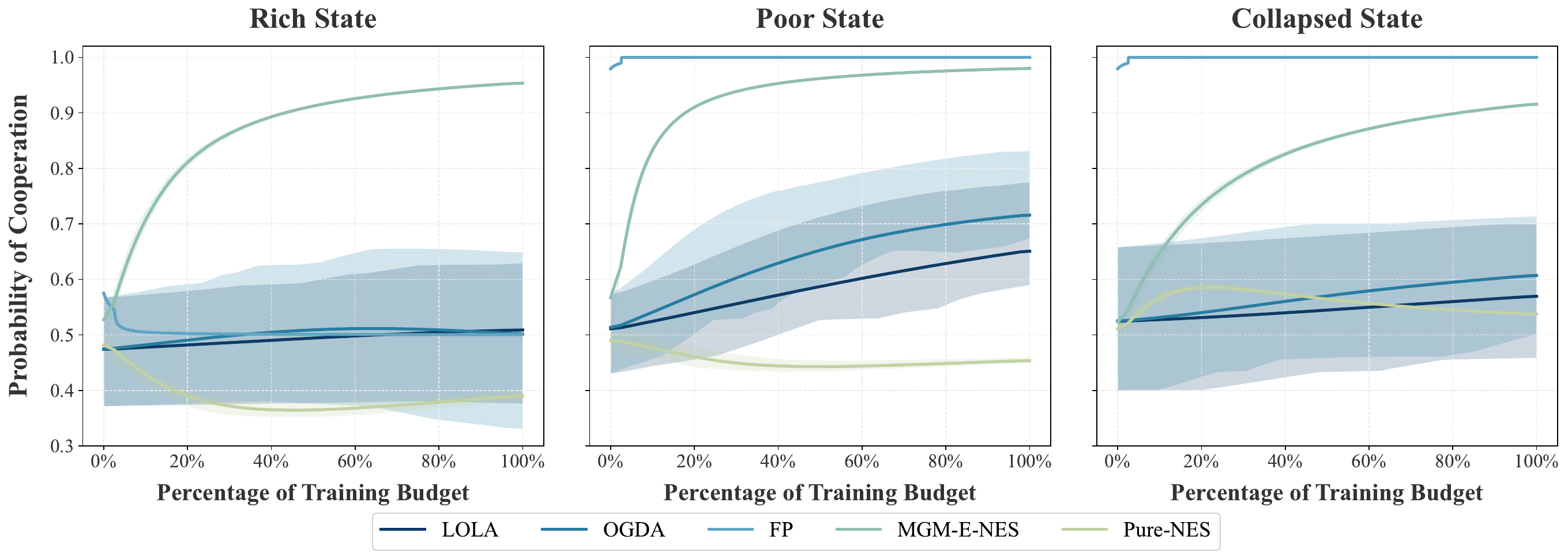}
    
    \caption{State-conditioned cooperation probability in the Markov resource game over training rollouts (x-axis; shown as percentage of the total rollout budget). Panels report cooperation in \textsc{Rich}, \textsc{Poor}, and \textsc{Collapsed} states, respectively. Curves show the mean over 30 random seeds; shaded regions indicate the interquartile range (25th–75th percentiles). MGM-E-NES is run with the same policy parameterization and hyperparameters as in the coordination-game setting (no task-specific retuning).}
    \label{fig:resource_game_results}
\end{figure}

Figure~\ref{fig:resource_game_results} reports the cooperation probability conditioned on each resource state. A consistent trend is that \textbf{FP attains the highest cooperation in \textsc{Poor} and \textsc{Collapsed}}, quickly approaching near-deterministic cooperation. This aligns with FP’s inductive bias: iteratively best-responding to the empirical opponent distribution can rapidly lock the population into a cooperative convention when cooperation is locally self-reinforcing. We report state-conditioned cooperation, which does not directly measure state visitation; welfare is determined by the induced stationary distribution and stage payoffs. 

\textbf{MGM-E-NES, by contrast, stands out for cross-state consistency and stability.} Although it does not always maximize cooperation in \textsc{Poor}/\textsc{Collapsed} relative to FP, it sustains a \textbf{substantially higher cooperation rate in \textsc{Rich}}, where several baselines remain near chance-level cooperation or drift toward less cooperative behavior. Since \textsc{Rich} is precisely the regime in which short-term incentives can make unilateral defection attractive, persistent cooperation there suggests that MGM-E-NES is not exploiting a state-specific shortcut, but instead learns a convention that is robust under opponent drift. The tighter interquartile bands further indicate a more reliable learning signal under noisy rollouts.

These state-wise curves also show that MGM-E-NES is \emph{not} “blindly cooperative”: it increases cooperation fastest in \textsc{Poor} (reaching \(0.8\) within \(\sim 10\%\) of the rollout budget), consistent with \textsc{Poor} acting as a high-leverage bottleneck in the resource dynamics. Joint cooperation both increases the chance of recovery to \textsc{Rich} \(\bigl(M_{\textsc{Poor}\rightarrow\textsc{Rich}}\propto p_{\textsc{Poor}}q_{\textsc{Poor}}\bigr)\) and avoids deterministic collapse under mutual defection \(\bigl(M_{\textsc{Poor}\rightarrow\textsc{Collapsed}}=(1-p_{\textsc{Poor}})(1-q_{\textsc{Poor}})\bigr)\). By comparison, \textsc{Rich} improves more gradually because it is already relatively stable (only \(DD\) triggers degradation), while \textsc{Collapsed} is slowest because recovery is sparse (only \(CC\) yields a small escape probability). Overall, the results support our claim that MGM’s cross-generational anchoring promotes \emph{stable} and \emph{welfare-aligned} behavior in mixed-motive Markov games, even when it does not maximize cooperation in every individual state.

\subsection{Overall Performance and Discussion}

The results summarized in Table~\ref{tab:ultimate_judgement} allow for a direct comparison of algorithmic performance across different game structures.

The data indicates that the performance of many baseline algorithms is highly dependent on the game type. For instance, methods like OGDA and FP demonstrate strong performance in specific domains, such as low-dimensional RPS. However, this performance does not consistently translate to other domains, like high-dimensional or coordination games.

In comparison, the results for MGM-E-NES show a different pattern. While it may not strictly outperform every baseline on every single metric, the data shows it is one of tested framework that maintains a high level of performance across the entire suite of evaluated tasks. This aligns with our primary design objective: to develop a general and adaptive solution.Notably, MGM-E-NES admits a largely shared hyperparameter setting across tasks. In particular, the 3-D RPS, coordination game, and Markov resource game can be run with an identical configuration. For higher-dimensional settings (e.g., 100D and 1000D RPS), we keep the MGM-E control-layer hyperparameters unchanged and adjust only the NES learning rate as a function of the action dimension. We further note that, in the main text, 3-D RPS uses a larger initial \(\sigma\) solely for clearer visualization; results with \(\sigma\) set identically to the coordination and resource games are provided in the Supplemental Material.

Furthermore, the contribution of individual components can be isolated. The strong convergence of PureNES in coordination games, where many baselines struggle, points to the effectiveness of the adaptive $\sigma$ mechanism. Our full ablation studies (see Supplemental Material) confirm that a standard NES without this mechanism fails to converge reliably in these scenarios. The consistent performance of the complete MGM-E-NES framework across all games thus illustrates the benefit of our top-level control approach for governing dynamics where other methods exhibit more specialized behavior.

\subsection{Evaluation Metrics and Wall-Clock Time}
\label{subsec:metrics}

\paragraph{Evaluation metric}
Throughout our experiments, we use the \textbf{Kullback--Leibler (KL) divergence} to quantify how close a learned strategy profile is to a target equilibrium strategy.
For two distributions $P$ and $Q$ over actions, the KL divergence is defined as
\begin{equation}
D_{\mathrm{KL}}(P \,\|\, Q) \;=\; \sum_{a} P(a)\log\frac{P(a)}{Q(a)} \, .
\end{equation}
In our setting, $P$ denotes the agent's (or joint) empirical strategy distribution and $Q$ denotes the target equilibrium distribution. Smaller values indicate closer alignment, and $D_{\mathrm{KL}}=0$ indicates an exact match, providing an interpretable measure of convergence toward stable outcomes~\cite{kullback1951}.

\paragraph{Wall-clock time}
To assess practical viability, Tables~\ref{tab:wallclock_rps} and~\ref{tab:wallclock_coord_markov} report approximate wall-clock time per training run (seconds, per seed) measured on a fixed hardware/software setup. All methods share the same implementation-level optimizations whenever applicable (e.g., batched evaluation and matrix sparsification for high-dimensional games).
We report two PSRO variants: \emph{PSRO (accelerated)} uses parallel rollout workers because a non-parallel implementation was prohibitively slow, whereas \emph{PSRO (fair)} is timed without parallelism but only for $10\%$ of the full evaluation budget (a full run is expected to take $\approx 10\times$ longer). Complete hardware/software specifications and timing configurations are provided in the supplementary material.

 
\begin{table}[t]
\centering
\caption{Approximate wall-clock time per training run (seconds, per seed) on RPS environments.}
\label{tab:wallclock_rps}
\setlength{\tabcolsep}{5pt}
\renewcommand{\arraystretch}{1.1}
\small
\begin{tabular}{lccc}
\toprule
\textbf{Algorithm} & \textbf{3D-RPS} & \textbf{100D-RPS} & \textbf{1000D-RPS} \\
\midrule
\textbf{MGM-E-NES (ours)} & \textbf{13} & \textbf{49} & \textbf{372} \\
PureNES & 6 & 35 & 150 \\
\midrule
PSRO (accelerated)\footnotemark[1] & N/A & N/A & 1800 \\
PSRO (fair)\footnotemark[2] & 100 & 800 & 3120 \\
\midrule
OGDA & 30 & 26 & 5 \\
LOLA & 50 & 80 & 10 \\
DPG  & 17 & 330 & 90 \\
Fictitious Play (FP) & 2 & 4 & 1 \\
\bottomrule
\end{tabular}

\footnotetext[1]{\textit{PSRO (accelerated)} uses parallel rollout workers because non-parallel PSRO was prohibitively slow in our settings.}
\footnotetext[2]{\textit{PSRO (fair)} is timed for only 10\% of the full evaluation budget; a full run would be approximately $10\times$ longer.}
\end{table}

\begin{table}[t]
\centering
\caption{Approximate wall-clock time per training run (seconds, per seed) on coordination and Markov-game environments.}
\label{tab:wallclock_coord_markov}
\setlength{\tabcolsep}{5pt}
\renewcommand{\arraystretch}{1.1}
\small
\begin{tabular}{lccc}
\toprule
\textbf{Algorithm} & \textbf{Stag Hunt} & \textbf{Battle of Sexes} & \textbf{Markov game} \\
\midrule
\textbf{MGM-E-NES (ours)} & \textbf{4} & \textbf{4} & \textbf{505} \\
PureNES & 2 & 2 & 253 \\
\midrule
PSRO (accelerated)\footnotemark[1] & N/A & N/A & N/A \\
PSRO (fair)\footnotemark[2] & 23 & 23 & 23 \\
\midrule
OGDA & 50 & 50 & 1 \\
LOLA & 4 & 5 & 3 \\
DPG  & 40 & 40 & N/A \\
Fictitious Play (FP) & 3 & 3 & 1 \\
\bottomrule
\end{tabular}
\end{table}

\section{Conclusion and Future Work}
\label{sec:conclusion}

This paper investigates two pervasive challenges in competitive coevolution: (i) evaluation non-stationarity induced by opponent drift, and (ii) unreliable progress signals under noisy rollouts. Together, they often lead to pathological evolutionary dynamics such as cycling, Red-Queen effects, and detachment. To address the central question---how to construct stable, cross-generation progress signals under black-box evaluation---we propose the \emph{Marker Gene Method} (MGM). Inspired by curriculum-learning principles, MGM introduces marker individuals that act as anchors, so that cross-generation comparisons are made against relatively stable references. Building on this idea, we further design the DWAM mechanism and explicit marker update criteria to suppress unreliable updates under noise and non-stationarity, thereby stabilizing selection pressure. For strictly competitive games, we provide supporting theoretical analysis and verify on RPS-type games that MGM can effectively mitigate unstable dynamics.

To reduce the manual sensitivity of MGM to the key threshold \(l\), we propose NGD-Div, which adapts \(l\) automatically by using a divergence measure around equilibrium as a proxy signal and optimizing it via natural gradient descent. Rather than directly prescribing an optimization direction, this mechanism modulates the triggering strength/frequency of marker updates, indirectly shaping the effective evolutionary gradient field and improving robustness and transferability under noisy evaluations.

We conduct systematic evaluations of MGM-E-NES on RPS-type games, coordination games (including Stag Hunt and Battle of the Sexes), and a resource-depletion Markov game. Across experiments, we keep the core hyperparameters of MGM's top-level control as consistent as possible, adjusting only dimension-dependent settings of the underlying NES when necessary (e.g., learning-rate scaling or initialization for different action dimensions). For tasks with comparable policy parameterization---3D RPS, the coordination games, and the resource-depletion setting---we further demonstrate that an identical hyperparameter configuration is viable (see Supplemental Material). Overall, the results indicate that MGM-E-NES substantially alleviates coevolutionary pathologies: in RPS-type experiments, this is reflected by reduced divergence and more stable training; in coordination and resource-depletion tasks, it manifests as a more stable, welfare-aligned optimization behavior. In particular, MGM-E-NES reliably converges to Pareto-superior Nash equilibria in Stag Hunt and Battle of the Sexes. In the Markov resource game, the learned behavior exhibits state-dependent cooperation aligned with long-horizon returns and social welfare, rather than indiscriminately maximizing cooperation in every state.

Several directions remain for future work. First, while our theoretical characterization builds on the SCG framework and specific assumptions, extending it to more general mixed-motive Markov games (including partial observability, asymmetry, and multi-agent generalizations) is an important next step. Second, MGM-E-NES can still be improved in sample efficiency; integrating it with parallelized evaluation, ERL-style pipelines, or more efficient black-box optimization techniques may reduce rollout cost. Third, MGM's governance mechanism is optimizer-agnostic: it can be combined with other black-box optimizers, or hybridized with gradient-based opponent modeling, to further improve stability and sample efficiency.

\section*{Data Availability Statement}
The anonymized dataset and code are available via OSF (view-only):
\url{https://osf.io/zbuk2/overview?view_only=bead0dc099804c658174ec180b11eb3c}.

\bibliographystyle{elsarticle-num-names} 
\bibliography{Bibliography}
\end{document}